\documentclass[conference]{IEEEtran}
\ifCLASSINFOpdf
\else
\fi
\usepackage{tikz}
\usepackage{graphicx}
\usepackage{dblfloatfix}  
\usepackage{svg}
\usepackage[ruled, lined, linesnumbered, commentsnumbered, longend]{algorithm2e}
\usepackage{amsfonts}
\usepackage{amsmath}
\usepackage{amsbsy}
\usepackage{bm}
\usepackage{cases} 
\usepackage{mathtools}
\usepackage{hyperref}
\usepackage{multirow}
\usepackage{booktabs}
\usepackage{comment}
\usepackage{amssymb}
\usepackage{xcolor}
\usepackage{xspace}

\usepackage[normalem]{ulem}

\SetKw{Continue}{continue}
\SetKw{Break}{break}
\SetKw{Terminate}{terminate}

\newcommand{\GE }{\text{GradEstimation}\xspace}
\newcommand{\RSBCD }{BlockDescent\xspace}
\newcommand{\rambo }{RamBoAttack\xspace}
\newcommand{\rambos }{RamBoAttacks\xspace}
\newcommand{\hardset }{\textit{hard-set}\xspace}
\newcommand{\hardsets }{\textit{hard-sets}\xspace}
\newcommand{\easyset }{\textit{non-hard set}\xspace}
\newcommand{\hard }{\textit{hard}\xspace}
\newcommand{\easy }{\textit{non-hard}\xspace}
\newcommand{\hardupper }{\textit{Hard}\xspace}
\newcommand{\easyupper }{\textit{Non-hard}\xspace}
\newcommand{\etal}{\textit{et al.}}

\newcommand{\ie}{\textit{i.e.}}

\newcommand\sbullet[1][.5]{\mathbin{\vcenter{\hbox{\scalebox{#1}{$\bullet$}}}}}

\definecolor{myyellow}{rgb}{0.9,0.8,0.05}

\usepackage{tcolorbox}
\newcommand{\rqq}[1]{
\begin{center}
\begin{tcolorbox}[width=\columnwidth, 
boxsep=0pt,
boxrule=0pt,
toprule=0pt,
colback=blue!5,
arc=5pt,
auto outer arc]
\textit{#1}
\end{tcolorbox}
\end{center}
}

\begin{document}
%
\title{\Large \bf RamBoAttack: A Robust Query Efficient Deep Neural Network Decision Exploit} 


\author{\IEEEauthorblockN{Viet Quoc Vo}
\IEEEauthorblockA{The University Of Adelaide\\
viet.vo@adelaide.edu.au}
\and
\IEEEauthorblockN{Ehsan Abbasnejad}
\IEEEauthorblockA{The University Of Adelaide\\
ehsan.abbasnejad@adelaide.edu.au}
\and
\IEEEauthorblockN{Damith C. Ranasinghe}
\IEEEauthorblockA{The University Of Adelaide\\
damith.ranasinghe@adelaide.edu.au}
}


\IEEEoverridecommandlockouts
\makeatletter\def\@IEEEpubidpullup{3\baselineskip}\makeatother
\IEEEpubid{\parbox{\columnwidth}{
    Network and Distributed Systems Security (NDSS) Symposium 2022\\
    27 February - 3 March 2022\\
    ISBN 1-891562-66-5\\
    https://dx.doi.org/10.14722/ndss.2022.24200\\
    www.ndss-symposium.org
}
\hspace{\columnsep}\makebox[\columnwidth]{}}

\maketitle

\begin{abstract}
Machine learning models are critically susceptible to evasion attacks from adversarial examples. Generally, adversarial examples---modified inputs deceptively similar to the original input---are constructed under whitebox access settings by adversaries with full access to the model. However, recent attacks have shown a remarkable reduction in the number of queries to craft adversarial examples using blackbox attacks. Particularly alarming is the now, \textit{practical}, ability to exploit simply the classification decision (\textit{hard label only}) from a trained model's \textit{access interface} provided by a growing number of Machine Learning as a Service (MLaaS) providers---including Google, Microsoft, IBM---and used by a plethora of applications incorporating these models. An adversary's ability to exploit \textit{only} the predicted label from a model-query to craft adversarial examples is distinguished as a \textit{decision-based} attack. 

In our study, we first deep-dive into recent state-of-the-art decision-based attacks in ICLR and S\&P to highlight the costly nature of discovering low distortion adversarial employing approximate gradient estimation methods. We develop a \textit{robust} class of \textit{query efficient} attacks capable of avoiding entrapment in a local minimum and misdirection from noisy gradients seen in gradient estimation methods. The attack method we propose, \textit{RamBoAttack}, exploits the notion of Randomized Block Coordinate Descent to explore the hidden classifier manifold, targeting perturbations to manipulate only localized input features to address the issues of gradient estimation methods. Importantly, the \textit{RamBoAttack} is demonstrably more robust to the different sample inputs available to an adversary and/or the targeted class. Overall, for a given target class, \textit{RamBoAttack} is demonstrated to be more robust at achieving a lower distortion within a given query budget. We curate our extensive results using the large-scale high resolution \texttt{ImageNet} dataset and open-source our attack, test samples and artifacts on \texttt{GitHub}.
\end{abstract}

\section{Introduction}

Demonstrations of super human performance from Machine Learning (ML) models, particularly Deep Neural Networks (DNNs), are leading to the industrialization of Machine Learning exemplified by self-driving cars \cite{Chen2015} and MLaaS from a plethora of providers, including IBM Watson Visual Recognition \cite{IBMWatson}, Amazon Rekognition \cite{Amazon} or Microsoft’s Cognitive Services \cite{azureCS}. Now, at the cost-per-service level, any system can easily integrate \textit{intelligence} into applications. The increasingly inevitable, wide spread proliferation of machine learning in systems are creating the incentives and \textit{new} attack surfaces to exploit, for malevolent actors. 

\begin{figure}[ht]
    \begin{center}
        \includegraphics[scale=0.62]{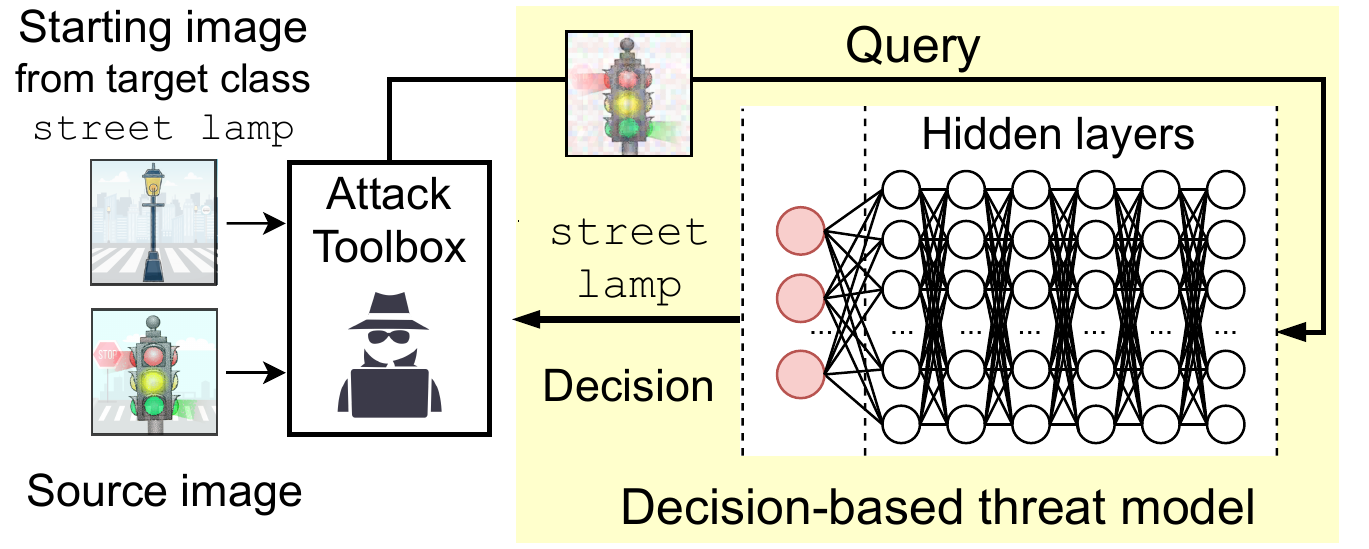}
        \caption{An illustration of blackbox attack in the severely restricted threat model of a decision-based attack. In a decision-based threat model, an adversary with a source image and starting image from target class, crafts a sample, query the model and observe the decision returned by the model.}
        \label{fig:hard-case1}
    \end{center}
    \vspace{-3mm}
\end{figure}

\vspace{1mm}
\noindent\textbf{Adversarial Attacks in White-box Settings.~}In particular, machine learning models are critically vulnerable to evasion attacks from carefully crafted adversarial examples. An adversary yields small perturbations, when added to an input, cause a failure---simply misclassifying the input in an \textit{untargeted attack} or hijacking the decision of a model to generate a decision pre-selected by the adversary~\cite{Szegedy2013} in a \textit{targeted attack}. Effective attack methods for generating adversarial examples in \textit{white-box} attacks, assuming full knowledge and access to the machine learning models, exist~\cite{Goodfellow2014, Papernot2016b, Madry2017, Carlini2017, Xu2019}.

\vspace{1mm}
\noindent\textbf{Adversarial Attacks in Blackbox Settings.~}In contrast, on commercial and industrial systems, 
an attacker has limited or no knowledge of model architecture, parameters or weights. Access may be limited to only full or partial output scores (treated as  a probability distribution). Chen et al. \cite{Chen2017} and Ilyas et al.~\cite{Ilyas2018} proposed methods to exploit models revealing output scores to craft adversarial examples under so-called \textit{score-based attacks}.In the \textit{most} restricted threat model illustrated in Fig.~\ref{fig:hard-case1}, the information exposed to an attacker is limited to the \textit{hard-label} only---the most confident label predicted or \textit{decision}, for instance logo or landmark detection on Google Cloud Vision \cite{googlecloudvision}.

\vspace{1mm}

\noindent\textit{Adversarial attacks in such a decision-based scenario are the most restrictive and challenging attack setting given the severely limited access to information, but, these settings present a realistic and pragmatic threat model.} 

\vspace{1mm}
\noindent\textbf{Decision-Based (Hard-Label) Adversarial Attacks.~}Recent studies demonstrated the practicability of blackbox attacks under the highly restrictive decision-based attack setting relying \textit{solely} on the label obtained from model queries. The Boundary Attack of Brendel et al.~\cite{Brendel2018} in ICLR demonstrated the feasibility of an attack and obtained adversarial examples comparable with state-of-the-art white-box attack methods in both \textit{targeted} and \textit{untargeted} scenarios. 

For a realistic attack, achieving attack success with a limited query budget is important because: i) MLaaS providers limit the rate of queries to their services; ii) throttling at a service provider limits large-scale attacks; and iii) a provider can employ methods to recognize a large number of rapid queries in succession with similar inputs to detect malicious activity and thwart query inefficient attacks. Furthermore, from both an attacker perspective and a defense perspective, reducing the number of queries \textit{reduces the cost of mounting the attack} as well the time for evaluating the model and potential defenses\footnote{For example, we consumed over 1,700 hours on two dedicated modern GPUs with 48~GB memory to curate the results in our study.}.

\vspace{-2mm}
\subsection{Our Motivation and Attack Focus}
Recent studies formulated the decision-based attack as an optimization problem to propose algorithms based on gradient estimation methods~\cite{Cheng2020,Chen2020} and demonstrated attacks with significantly reduced number of queries. However , the existing attacks suffer from the  following problems: 
\begin{itemize}
    \item \textit{Entrapment in a local minima.}In gradient estimation methods, as eluded to by Cheng \etal~\cite{Cheng2020}, the search for an adversarial example can experience an entrapment problem in a local minimum where extra queries expended by the attacker fails to achieve a lower distortion adversarial example. 
    
    \item\textit{Unreliabiilty of gradient estimations.~} Further, as the magnitude of estimated gradients diminish on approach to a local  minima  or a  plateau, the estimated gradients may become noisy and susceptible to misdirection.
    
    \item \textit{Sensitivity to  the   starting  image.~}Then, intuitively, we can expect that the initialization of optimization frameworks with an \textit{available} or intended starting image, a \textit{necessity} in decision-based attacks, to hinder an attacker from reaching an imperceptible adversarial example. But, there is no known method to determine \textit{a good starting image prior to an attack}. Thus, the success of an attack can be expected to be sensitive to the available starting image; an attempt to discover a better starting image or target class through \textit{trial and error} can not only lead to detection and discovery by effectively increasing the numbers of queries needed, but also  limit the scope of the attack by reducing the number of classes that can be targeted.
\end{itemize}

\vspace{1mm}
\noindent\textit{In general, developing decision-based attacks poses a challenging optimization problem because only binary information from output labels are available to us from the target model as opposed to output values from a function}.

\vspace{1mm}
Therefore, we seek to understand the fragility of gradient estimation methods and develop a more robust and query efficient attack. Consequently, we expend our efforts to answer the following research questions (RQ) .
\vspace{-2mm}
\rqq{RQ1: How can we assess the robustness of decision-based blackbox attacks to understand their fragility? (Section \ref{sec:SOTA approach})} 
\vspace{-4mm}
\rqq{RQ2: What is the impact of the source and starting target class images accessible to an adversary on the success of an attack? (Section \ref{observations} \& extensive results in Section~\ref{sec:sensitivity-to-starting-image})}
\vspace{-4mm}
\rqq{RQ3: How can an adversary construct a robust and query efficient attack for achieving low distortion adversarial examples for any starting image from the targeted class and avoid the pitfalls of gradient estimation based attack methods? (Section~\ref{sec:attack-approach}~\&~\ref{sec:evaluations})}
\vspace{-4mm}

\subsection{Our Contributions}
In our effort to: i)~address the research questions; ii)~better understand and assess the vulnerabilities of DNNs to adversarial attacks in the pragmatic decision-based threat model; and iii)~explore more robust attack methods, we summarize our contributions below: 

\begin{enumerate} 
    \item Our study presents the \textit{first} systematic investigation of  state-of-the-art decision-based attacks to understand their robustness. Through  extensive experiments, we highlight the problem of \hard cases where attackers struggle to flip the prediction of images towards a chosen target class, even with increasing query budgets--see Fig.~\ref{fig:hard-case matrix comparison}. 
    As summarized in Table~\ref{apd-experiment time summary}, we expend over 1800 computation hours with 2 GPUs to curate results. 
    
    \item Motivated by our findings, \textit{we propose a new attack}---\rambo---that is demonstrably more robust. We propose a search algorithm analogous to Randomized Block Coordinate Descent---\RSBCD---to address the entrapment problem where gradient estimation fails to provide a useful direction to descend and propose to combine \RSBCD with gradient estimation frameworks to attain query efficiency. In contrast to existing approaches, \RSBCD focuses on altering local regions of the input commensurate with the filter sizes employed by DNNs to forge adversarial examples. 
   
    \item We provide new insights into query efficient mechanisms for crafting adversarial perturbation to attack DNNs. Our decision-based blackbox attack method relying on localized alterations to inputs discovers effective adversarial perturbations attempting to exploit the model's reliance on salient features of the target class to correctly classify an input to a target label in the \hard cases. We illustrate clear correlations between perturbations found and added to inputs, and salient regions on target class images with the aid of a visual explanation tool. 
    
    \item Overall, \rambo is a more robust and query efficient approach for generating an adversarial example of high attack success rate compared to existing counterparts. Importantly, our attack method is \textit{significantly less impacted by a starting image} from a target class accessible to an adversary.
    
    \item We recognize the need for reliable and reproducible attack evaluation strategies and introduce two evaluation protocols applied across \texttt{CIFAR10} and \texttt{ImageNet}. We \textit{release the dataset constructed through our extensive study to support future benchmarking of blackbox attacks} under a decision-based setting\footnote{https://ramboattack.github.io/}. 

\end{enumerate}
\section{Decision-Based Attacks}
In this section, we: i)~formalize an adversarial attack as an optimization problem; ii)~revisit current state-of-the-art methods; and iii)~analyze the results to present our intuitions into the state-of-the-art attacks based on our observations.

\vspace{-2mm}
\subsection{Adversarial Threat Model}\label{sec:adv-threat-model}
We adopt the threat model proposed in prior works~\cite{Chen2020,Cheng2019,Brendel2018}. Under the decision-based blackbox setting, adversaries have no prior knowledge such as model architecture or parameters but have limited access to the output of a victim model---\textit{the model's decision} as illustrated in Fig.~\ref{fig:hard-case1}. Furthermore, an adversary can make numerous queries to a victim's  machine learning model via an access interface and receive the model's decision. The adversary must have \textit{at least one image from a target class that is classified correctly by the victim model if the adversary aims to carry out a targeted attack}. This image is the \textit{starting image} used to initialize the attack. The adversary also holds at least one image from a \textit{source class} correctly classified by the model. The \textit{objective} of the adversary is to discover the minimum (imperceptible) perturbation---quantitatively measured by the common distortion measure adopted in recent studies---to flip the decision for the source image to the targeted class using the minimum number of queries to the model.

\subsection{Problem Formulation}
Given a source image $\boldsymbol{x} \in \mathbb{R}^{C \times W \times H}$ its ground truth label $y$ from the label set $\mathcal{Y} = \{1, 2, \cdots, K\}$, where $K$ denote the number of classes, $C$, $W$ and $H$ denotes the number of channels, width and height of an image, respectively. Given a pre-trained multi-class classification model $f: \boldsymbol{x} \rightarrow y$ so that $f(\boldsymbol{x}) = y$, in a targeted attack, an adversary aims to modify an input $\boldsymbol{x}$ to craft an optimal adversarial example $\boldsymbol{x^{*}} \in \mathbb{R}^{C \times W \times H}$ that is classified as the class label desired by the adversary when used as an input for the victim model. In an untargeted attack, an adversary manipulates input $\boldsymbol{x}$ to change the decision of a classifier to any class label other than its ground-truth label. To simplify the descriptions, we refer to the desired class label as the \textit{target class} while the class of the input $\boldsymbol{x}$ is called the \textit{source class}.
\begin{figure*}[ht]
        \begin{center}
          \includegraphics[scale=0.26]{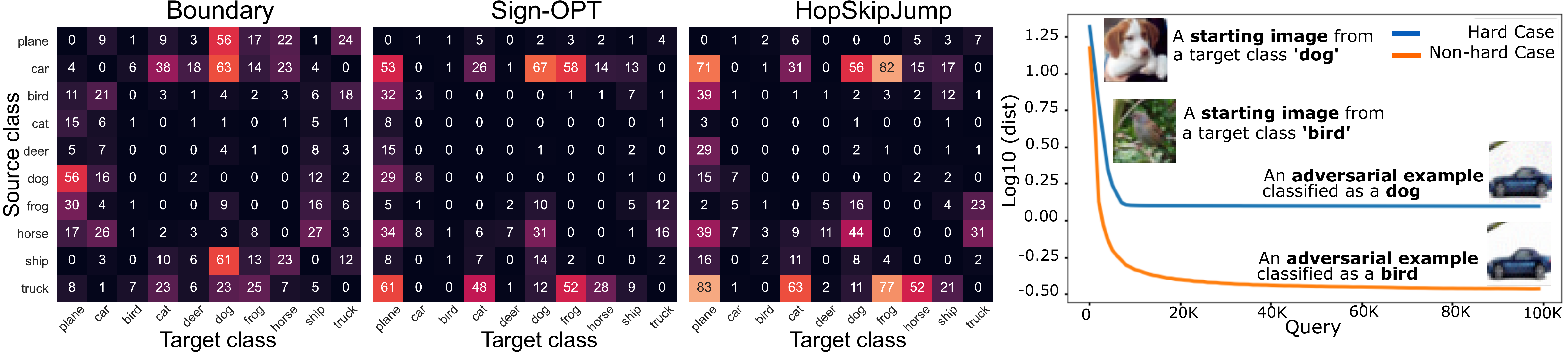}
          \caption{(Left) The number of \hard cases from \texttt{CIFAR10} found for Boundary Attack (BA), Sign-OPT and HopSkipJump categorized by different pairs of a source and target class (starting image) at a distortion threshold = 0.9 and a query budget of 50,000. (Right) The line chart illustrates a significant difference between a \hard versus \easy case---\textit{interestingly increasing the query budget to even 100,000 does not yield a lower distortion adversarial example}.}
          \label{fig:hard-case matrix comparison}
        \end{center}
        \vspace{-3mm}
\end{figure*}

\vspace{1mm}
\noindent\textbf{Measuring Distortion.~} $l_2$-norm is widely adopted, in \textit{all of the recent works} as  in~\cite{Brendel2018,Brunner2019,Cheng2019b,Cheng2020,Chen2017}, to measure the distortion and similarity between a generated adversarial example and the source sample. Therefore, in this paper, our approach focuses on $l_2$-norm. Then, let $D(\boldsymbol{x},\boldsymbol{x^{*}})$ be the $l_2$-distance that measures the similarity between $\boldsymbol{x}$ and $\boldsymbol{x^{*}}$.  

\vspace{1mm}
\noindent\textbf{Optimization Problem.~}The main aim of adversarial attacks is to minimize the distortion measured by $D$ while ensuring the perturbed input data is classified as a target class---for a targeted attack---or non-source class---for an untargeted attack. Therefore, an adversarial attack can be formulated as a constrained optimization problem:

\begin{equation}\label{eq:1}
\begin{aligned}
\min_{x^{*}} \quad & D(\boldsymbol{x},\boldsymbol{x^{*}}) \\ 
\textrm{s.t.} \quad & \mathcal{C}(f(\boldsymbol{x^{*}})) = 1, \\
\quad & \boldsymbol{x},\boldsymbol{x^{*}} \in [0,1]^{C \times W \times H},
\end{aligned}
\end{equation}

Here, $\mathcal{C}(f(\boldsymbol{x^{*}}))$ is an adversarial criterion that takes the value 1 if the attack requirement is satisfied and 0 otherwise. This requirement is satisfied if $f(\boldsymbol{x^{*}}) \neq y$ for an untargeted attack or $ f(\boldsymbol{x^{*}}) = y^*$  for a targeted attack (i.e. for the instance $\boldsymbol{x^{*}}$ to be misclassified as targeted class label $y^*$).

\subsection{Understanding Robustness} \label{sec:SOTA approach}
The two current query efficient attack methods employ gradient approximation frameworks, whilst the earlier method relied on a stochastic approach. We briefly summarize these methods before delving into our systematic study to understand their robustness.

\noindent\textbf{Random Walk along a Decision Boundary.} The first attack under a decision-based threat model proposed by Brendel et al. \cite{Brendel2018} initialized an image in a target class and in each iteration, sampled a perturbation from a Gaussian distribution and projected the perturbation onto a sphere around a source image.  If this perturbation yields an adversarial example, the attack makes a small movement towards the source image and repeats these steps until the decision boundary is reached. Subsequently, by traveling along the decision boundary based on sampling, projecting and moving towards the source image, the adversarial example is refined until an adversarial example with a desirable distortion is discovered.

\noindent\textbf{Optimization Frameworks.~}In the absence of a means for computing the gradient for solving \eqref{eq:1}, the attacks in~\cite{Cheng2019b} and~\cite{Cheng2020} attempt to solve the optimization problem using methods to estimate the gradient. Cheng et al.~\cite{Cheng2019b} samples directions from a Gaussian distribution and applies a zeroth-order gradient estimation method in their OPT-attack, then Cheng et al.~\cite{Cheng2020} leveraged their former optimization framework and proposed a  zeroth-order optimization algorithm called Sign-OPT that is much faster to converge. Chen et al. \cite{Chen2020} introduced a different optimization framework named HopSkipJumpAttack using a Monte Carlo method to find the approximate gradient direction to descend. 

\noindent\textbf{Evaluating Robustness.~}To understand the robustness of recent attack methods and illustrate the costly nature of discovering low distortion adversarial examples with these attacks, we propose an \textit{exhaustive but tractable} experiment using the relatively small number of classes albeit with a significantly large validation set offered by \texttt{CIFAR10} dataset. The protocol for assessing robustness of each state-of-the-art method described is carefully described in Appendix \ref{apd-Robustness Evaluation Protocol}. 

\noindent\textbf{Hard Cases.~}Empirically, we define a \hard case as a pair of source and starting images---the starting image is from a given target class---where a given decision-based attack fails to yield an adversarial example with a distortion below a desirable threshold using a set query budget. 

\subsection{Observations from Assessing Attacks} \label{observations}
\begin{figure*}[ht]
    \begin{center}
        \includegraphics[scale=0.34]{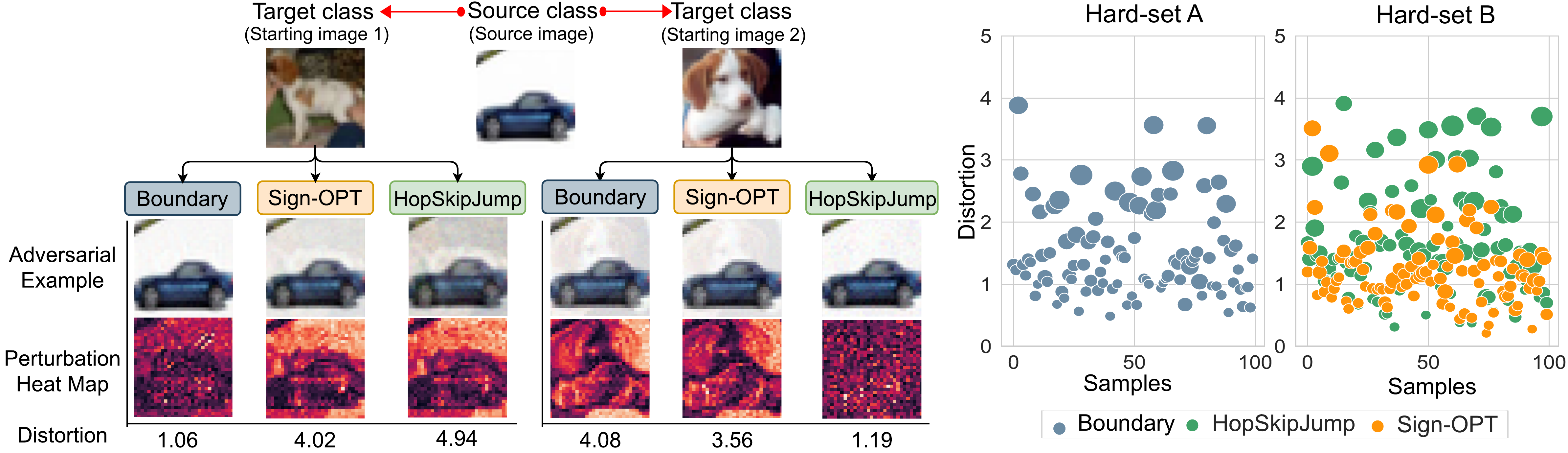}
        \caption{(Left) Consider an attack to discover an adversarial example for the source image of class \texttt{car} such that the car is predicted as belonging to the target class \texttt{dog}. We demonstrate the very different results an adversary can obtain based on the availability of a target class \textit{image1} and \textit{image2}. The attacker initializes the attack for Boundary, starting \textit{image1} is a better initialization. In contrast, Sign-OPT and HopSkipJump discover better adversarial examples if they are initialized with starting \textit{image2}. (Right) The scatter plot illustrates this attack scenario with 100 samples randomly selected from their own \hard set. The $y$-axis denotes the average distortion and the size of each bubble denotes the variation in distortion for each source image with respect to 10 different starting images from hard targets. It shows that all these methods are highly dependent on a starting image in \hard cases.
        }
        \label{fig:hard-case for different seed}
    \end{center}
    \vspace{-5mm}
\end{figure*}

We make the following observations from our results summarized in Figures~\ref{fig:hard-case matrix comparison}~and~\ref{fig:hard-case for different seed}.

\begin{figure*}[!b]
    \begin{center}

        \includegraphics[scale=0.36]{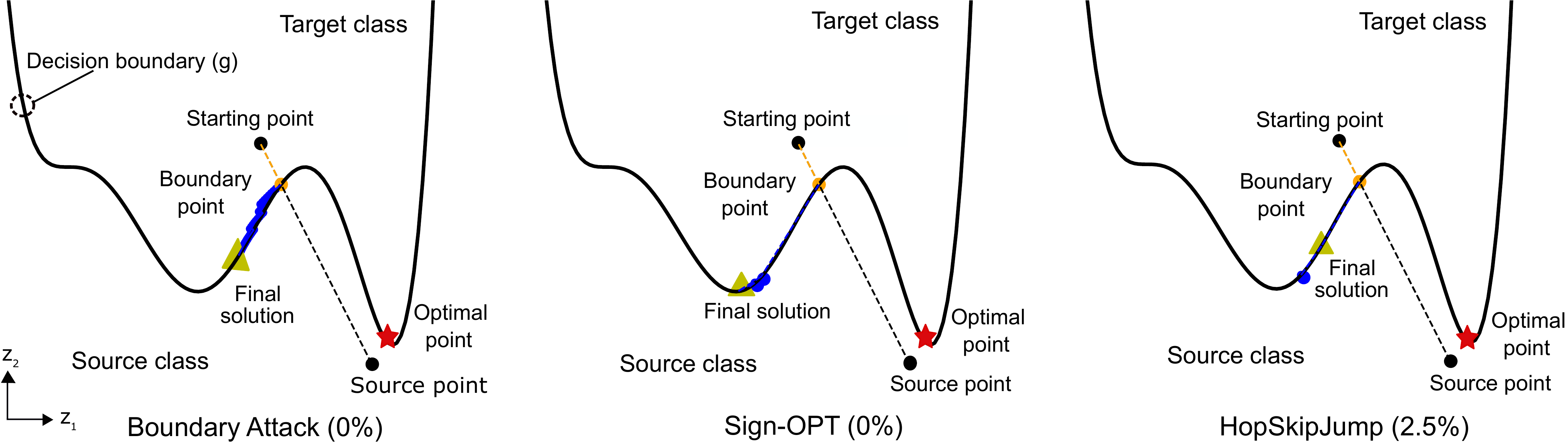}
        \caption{\textbf{2D ($z_1$ and $z_2$) Input Space Example.} An illustration of the execution of the three different decision based attack methods (Boundary, Sign-OPT and HopSkipJump) to attack a toy model employing 2D inputs. The attacks result in different final solutions denoted by a yellow triangle (${\color{myyellow}\blacktriangle}$). We executed the algorithms 100,000 times; both Boundary Attack and Sign-OPT failed to find the global minimum (the Optimal Point \textit{closest} to the Source point) and HopSkipJump only found the global minimum 2.5\% of the time. This illustrates the \textit{problem faced by current attack methods} when attacking a machine learning model \textit{whose decision boundary in the input space is multi-dimensional and highly complex} for realistic and practical image inputs.}
        \label{fig:toy-model}
    \end{center}
    \vspace{-3mm}
\end{figure*}


\vspace{1mm}
\noindent\textbf{Observation 1: Hard Cases.} \textit{In decision-based attacks, specific classes and/or samples from classes are more difficult to attack than others.} 
As illustrated in Fig.~\ref{fig:hard-case matrix comparison} (left), the current attack methods are not uniformly effective against all pairs of source and starting images from target classes. 

Interestingly, any of the gradient estimation methods can approximate the true gradient given enough queries (or samples) to the target model. However, solutions can become entrapped in various local minima. Further, approaching a local minimum or a plateau can considerably undermine the quality of that approximation; for instance, estimated gradients may become noisy when the gradient magnitude diminishes whilst approaching a local minimum. As shown in Fig.~\ref{fig:hard-case matrix comparison} (right), even with 100K queries, the solutions based on the gradient direction estimation methods do not improve the distortion of the adversarial sample for the \texttt{car} classified as a \texttt{dog} (Hard case).

\vspace{1mm}
\noindent\textbf{Observation 2: Attack initialization.~} \textit{An attack algorithm's ability to find a low distortion adversarial example with a given query budget is dependent on the starting image from a target class selected for initializing the attack algorithm.} Interestingly, Chen et al. \cite{Chen2020} in their S\&P2020 paper briefly noted the potential for an algorithm to get trapped in a bad local minimum based on the starting image used to initialize an attack. Our systematic study confirms this intuition.

In this case, the achievable distortion of an adversarial example is highly dependent on the starting image and the behavior of the algorithm. This observation is illustrated by comparing the results of starting \textit{image~1} with \textit{image~2} for different attack methods in Fig. \ref{fig:hard-case for different seed} and by 100 samples randomly selected from the \hard set of each method---see Section \ref{sec:benchmark on hard and non hard set} and \ref{sec:sensitivity-to-starting-image} for more details. \textit{The results demonstrates the dependence of attack success on the starting image accessible to an adversary}. 

\textit{Currently, there is no effective initialization method to determine a good starting image}, prior to mounting an attack. Therefore, developing robust attack that is less sensitive to the choice of starting image remains an open challenge. 

\vspace{1mm}
\noindent\textbf{Observation 3: Perturbation Region.} 
\textit{Current attack approaches aim to perturb the whole image to traverse the decision boundary to find an adversarial example with minimum distortion.} In other words, these methods always manipulate the whole image at a time and result in perturbations that is spread over the entire image as illustrated by perturbation heat maps in Fig. \ref{fig:hard-case for different seed}. Another interesting remark drawn from these figures is that the main features (for example edges) of the starting image remains super-imposed in an adversarial example. However, most of the state-of-the-art classifiers in computer vision utilize convolutional filters to extract local patterns in an image; further, visual explanation tools demonstrate the reliance of classifiers on key salient features of an image. \textit{Therefore, whether an attack could achieve a lower distortion adversarial example by targeting the filter operation over local features in contrast to manipulation of the whole image remains an interesting direction to explore}.

\subsection{An Intuition into Attack Methods} \label{intuition}

To understand and illustrate the underlying cause of the first two observations, we use Boundary attack (BA) \cite{Brendel2018}, Sign-OPT \cite{Cheng2020} and HopSkipJump~\cite{Chen2020} to attack a Toy model. The \textit{decision boundary} of the Toy model in a 2D \textit{input space} illustrates a constraint of the optimization problem in~\eqref{eq:1}. This decision boundary is represented by $g(z_1,z_2) = (z_1-2)(z_1-1)^2(z_1+1)^3-z_2 = 0$ where $z_1$ and $z_2$ denote two coordinates of a point such as a starting point or a source point as illustrated in Fig. \ref{fig:toy-model}. A point above the boundary is classified as in target class; otherwise, it belongs to the source class. The black dot ($\sbullet[0.85]$) \textit{source point} denotes a source class example whilst black dot ($\sbullet[0.85]$) \textit{starting point} denotes a starting target class example. All three methods are initialized with the same starting point, we then employ the attacks to search for an adversarial point within the target class and closest to the source point; alternatively, we aim to solve the optimization problem in~\eqref{eq:1}, where the objective is to minimize the $l_2$ distance to the source point subject to the constraint imposed by the decision boundary, using these attack algorithms. 

\begin{figure*}[ht]
\begin{center}
\includegraphics[scale=1]{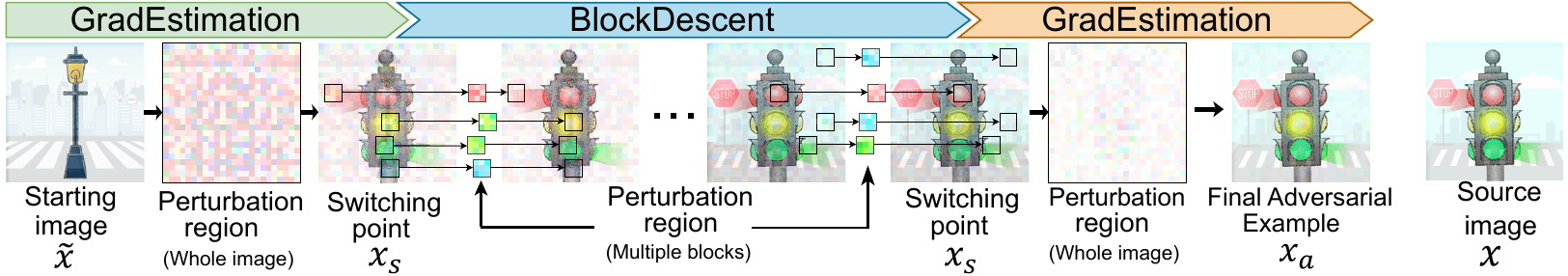}
\caption{A \textit{\textbf{pictorial illustration}} of \rambo to craft an adversarial example. In a targeted attack, the first component (GradEstimation) initializes an attack with a starting image from a target class (e.g. we use a clip art \texttt{street lamp} for illustration) and then manipulates this image to search for adversarial examples that looks like an image from source class e.g \texttt{traffic light}. The attack switches to the second component, \RSBCD, when it reaches its own local minimum. \RSBCD helps to redirect away from that local minimum by manipulating multiple blocks---or making local changes to the current adversarial example. Subsequently, the adversarial example crafted by \RSBCD is refined by the third component (GradEstimation).}

\label{fig:overal process}
\end{center}
\vspace{-3mm}
\end{figure*}

Fig.~\ref{fig:toy-model} illustrates several intermediate adversarial example points denoted by blue dots and a final adversarial example achieved by each method denoted by a yellow triangle for one example attack execution. Given the stochastic nature of the algorithms, we execute each attack 100,000 times with different random seeds. All of the methods, except HopSkipJump fails to find the optimal solution---global minimum---and HopSkipJump only managed to reach the optimal solution in 2.5 $\%$ of the attempts. As illustrated in Fig.~\ref{fig:toy-model}, the approximate gradient appears to be noisy and the methods traverses the decision boundary in an incorrect direction towards the local minimum rather than the global minimum. Although not illustrated here, changing the starting coordinate can lead all of these methods to discover the global minimum. 

\section{Proposed Attack Framework}\label{sec:attack-approach}

We observe that: i) gradient estimation methods in attacks face an entrapment problem in a highly complex loss landscape; ii) current attacks focus on altering all of the coordinates of an image simultaneously to forge a perturbation; and iii) the success of current attacks are sensitive to the chosen or available starting image possessed by an adversary. 

We propose an analogous Randomized Block Coordinate Descent method---named \RSBCD---that aims to manipulate local features and target convolutional filter outputs by modifying values of coordinates in a square-block region and in different color channels with targeted perturbations. We propose localized changes to affect convolutional filter outputs and pixel values as a means of impacting on salient features and may be even mimic salient features of the target. This leads to potential redirection and escape from entrapment in a bad local minimum with minimal but effective changes to the image to mislead the classifier. In other words, we propose taking a direct path along some coordinates towards a source image whilst retaining the target class label to prevent the problem encountered by gradient estimation methods---entrapment in a local minimum as shown in Fig.\ref{fig:toy-model}.

\begin{figure}[ht]
\begin{center}
\includegraphics[scale=1.1]{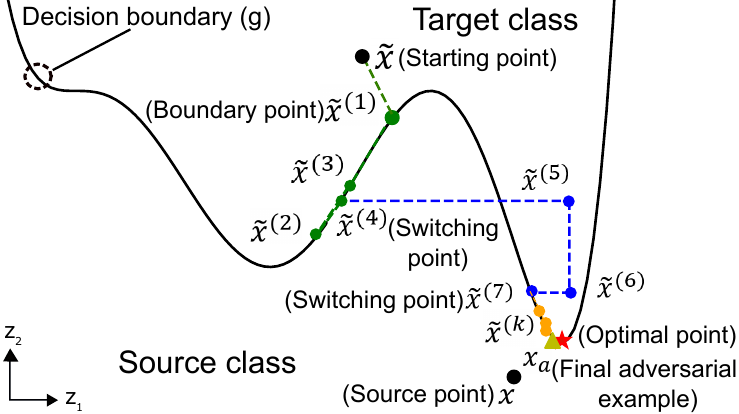}
\caption{\textbf{2D ($z_1$ and $z_2$) Input Space Example.} An illustration of our \rambo against the toy model in Fig.~\ref{fig:toy-model}.
If the first gradient estimation method---\GE in Algorithm~\ref{algo:hybrid_method}---leads to entrapment in a local minimum---denoted by $\boldsymbol{\Tilde{x}}^{(1)},\cdots,~\boldsymbol{\Tilde{x}}^{(4)}$ at the start---there is no effective mechanism to escape. However \RSBCD moves away from the local minimum. This is illustrated by $\boldsymbol{\Tilde{x}}^{(5)},\cdots,~\boldsymbol{\Tilde{x}}^{(7)}$ when the number of modified coordinates is one in the 2D input space. Subsequently, the third component applying a gradient estimation method searches for a better adversarial example $\boldsymbol{\Tilde{x}}^{(k)}$ in the neighborhood region and reaches the nearly optimal solution $\boldsymbol{x}_\text{a}$. In contrast to results in Fig.~\ref{fig:toy-model}, when evaluating \rambo over 100,000 runs against the Toy model, we observed our attack to \textbf{\textit{always}} find the optimal or near optimal solution.}

\label{fig:toy example - hybrid}
\end{center}
\vspace{-3mm}
\end{figure}

Further, when employing gradient estimation methods, the gradient values decrease as we move closer to the source image leading to increasingly larger number of perturbations needed to converge. This issue is exacerbated if there is a plateau in the decision boundary; now the gradient estimation methods are  as effective as a random search. We conjecture that the \hard cases are examples of where the gradient of the distortions are generally small and, thus, leads to a local optima. However, we observe that the gradient estimation methods are effective in two cases: \textbf{(a)} initial stages of optimizing Eq.~(\ref{eq:1}) or \textbf{(b)} at close proximity to the source image. In (a), the gradients are sufficiently large to be estimated effectively, and in (b) 
small changes and refinements (\ie~few perturbation iterations) facilitate a decent to the optimum. 

Consequently, we propose a new framework using gradient estimation for the initial descent---case (a)---supported by \RSBCD to escape entrapment and noisy gradient problems and refining the adversarial example supported by a gradient estimation based descent to forge a robust and query efficient attack. Importantly, \RSBCD is \textit{insensitive to the choice of starting images}, although it is effectively initialized with a gradient estimation, because \RSBCD manipulates blocks that causes a move away from the direction set by a starting images. The new framework we propose, \rambo, is illustrated in Fig.~\ref{fig:overal process}.

\vspace{1mm}
\noindent\textbf{Summary.~}\textit{Gradient estimation methods are fast but face the potential problem of getting trapped in a bad local minimum, particularly in \hard cases. \RSBCD, on the other hand, is slower---selecting to manipulate local regions---but is capable of tackling the problems faced by gradient estimation attacks. Therefore, we develop a hybrid framework called \rambo for query efficient decision-based attacks that can exploit the merits of both approaches. In particular, our derivative-free optimization method considers, for the first time, an approach to manipulate blocks of coordinates in the input image to influence the outcome of convolution operations used in deep neural networks as a means for misguiding a networks decision and generating adversarial examples with minimal manipulations.}

\subsection{Approach} \label{approach}
Our proposed attack thus comprises of \RSBCD and two components of gradient estimation---GradEstimation---as shown in Fig. \ref{fig:overal process} and described in Algorithm \ref{algo:hybrid_method}. The gradient estimation algorithms used by these two components can be the same or different from each other. 
When starting an attack, particularly in targeted setting, the first component is initialized with a starting image $\boldsymbol{\Tilde{x}}$ from a target class and approaches the decision boundary via a binary search---the first step in a gradient estimation method. We employ the gradient estimation method to search for adversarial examples until reaching its own local minimum. We call it a switching point $\boldsymbol{x_\text{s}}$ because from this point, gradient estimation method switches to \RSBCD. If the gradient estimation method is entrapped in local minimum, \RSBCD helps to move away from that local minimum. Subsequently, when local changes are insufficient, the attack switches to the third component to refine the adversarial example crafted by \RSBCD which is considered as the second switching point. This refinement aims to search for the final adversarial example $\boldsymbol{x}_\text{a}$ with a lower distortion. 

Fig. \ref{fig:toy example - hybrid} illustrates \rambo against the Toy model used in Section \ref{intuition} and demonstrates the effectiveness of the attack we propose. Particularly, the first gradient estimation approach searches for and reaches the adversarial examples $\boldsymbol{\Tilde{x}}^{(1)},~ \boldsymbol{\Tilde{x}}^{(2)},~ \boldsymbol{\Tilde{x}}^{(3)}$ at different steps towards approaching the source point but is stuck
at $\boldsymbol{\Tilde{x}}^{(4)}$ which is a local minimum of the objective function $D(\boldsymbol{x},\boldsymbol{x^{*}})$ subject to the constraint defined by the decision boundary $g(z_1,z_2)$. 
Henceforth, \RSBCD searches for next adversarial examples $\boldsymbol{\Tilde{x}}^{(5)}, \cdots, ~\boldsymbol{\Tilde{x}}^{(7)}$ by modifying one coordinate at a time---in this 2D example---by applying $\delta$ changes. Subsequently, the second gradient estimation method continues searching for adversarial examples $\boldsymbol{\Tilde{x}}^{(k)}$ in the neighborhood areas until reaching the near optimal $\boldsymbol{x}_\text{a}$. Most importantly, in contrast to experiment in Fig.~\ref{fig:toy-model} when evaluating \rambo over 100,000 attacks on the Toy model, our proposed attack always reached the optimal or near optimal solution.

\vspace{1mm}
\noindent\textbf{When to switch to \RSBCD?~} The gradient estimation methods are designed to work alone rather than with other methods. Therefore, we develop a sub-module \GE to call these methods and determine when to switch from a gradient estimation method to \RSBCD. Empirically, gradient estimation methods reach their local minimum when they cannot find any better adversarial example after several steps of searching. In practice, this can be determined by the distortion reduction rate $\Delta$ after every $T$ queries---a time frame to calculate $\Delta$. However, in gradient estimation methods, the number of queries per iteration is varied so we relax this by accumulating the number of queries after each iteration. Whenever it exceeds $T$, we compute $\Delta$ and if this distortion reduction rate is below a switching threshold $\epsilon_\text{s}$, it switches to \RSBCD (see Algorithm \ref{algo:gradestimation}).

\begin{algorithm}[t]
    \SetKwInOut{KwIn}{Input}
    \DontPrintSemicolon
   
    \KwIn{source image $\boldsymbol{x}$, starting image $\boldsymbol{\Tilde{x}}$, model $f$\;

    gradient estimation function $g_1,~g_2$, reduction scale $\lambda$,\;
    input dimensions $N$,square extension $n$,\;
    block number $m$, query number $T_1,T_2$}
      $\boldsymbol{x}_\text{s}\leftarrow $ \GE $(\boldsymbol{x},\boldsymbol{\Tilde{x}},f,g_1,T_1)$\;
    $\boldsymbol{x}_\text{s}\leftarrow$ \RSBCD$(\boldsymbol{x},\boldsymbol{x}_\text{s}, f, \lambda, N, n, m)$\;
    $\boldsymbol{x}_\text{a}\leftarrow $ \GE $(\boldsymbol{x},\boldsymbol{x}_\text{s},f,g_2,T_2)$\;
    \KwRet{$\boldsymbol{x}_\text{a}$}\;
    \caption{RamBoAttack}
    \label{algo:hybrid_method}
\end{algorithm}

\begin{algorithm}[t]
    \SetKwInOut{KwIn}{Input}
    \DontPrintSemicolon

    \KwIn{source image $\boldsymbol{x}$, switching image $\boldsymbol{{x}_\text{s}}$, model $f$\;
    \quad \quad \quad  gradient estimation function $g$, query number $T$}

        $n_\text{q} \leftarrow 0$, $switch \leftarrow False$\;
        $d \leftarrow D(\boldsymbol{x},\boldsymbol{x}')$\;
        \While{$not ~(switch)$}{
            $\boldsymbol{x}',~i\leftarrow g(f,\boldsymbol{x},\boldsymbol{x}')$\;
            $n_\text{q} \leftarrow n_\text{q} + i$\; 
            \uIf{$n_\text{q} ~> ~T$}{
                $\Delta \leftarrow d - D(\boldsymbol{x},\boldsymbol{x}')$\;
                $d \leftarrow D(\boldsymbol{x},\boldsymbol{x}')$\;
                $n_\text{q} \leftarrow 0$\;
                \uIf{$\Delta < \epsilon_\text{s}$}{
                    $switch \leftarrow True$\;}
                }
            }
        \KwRet{$\boldsymbol{x}'$}
    \caption{GradEstimation}
    \label{algo:gradestimation}

\end{algorithm}
\vspace{-2mm}
    
\begin{algorithm}
    \SetKwInOut{KwIn}{Input}
    \DontPrintSemicolon
    \KwIn{source image $\boldsymbol{x}$, switching image $\boldsymbol{x}_\text{s}$,model $f$\; 
    reduction scale $\lambda$, input dimension $N$\;
    square extension $n$, block number $m$}
    $k \leftarrow 0$, $n_\text{q} \leftarrow 0$, $switch \leftarrow False$\;
    $\delta \leftarrow P_\text{i}(|\boldsymbol{x} - \boldsymbol{x}_\text{s}|)$,
    $\boldsymbol{\Tilde{x}}^{(k)} \leftarrow \boldsymbol{x}_\text{s}$, $D_{n_\text{q}}\leftarrow D(\boldsymbol{x},\boldsymbol{\Tilde{x}}^{(k)})$\;
    \While{not (switch)}{
        $j \leftarrow 0$ \;
        \While {$j < ~N ~\textbf{and} ~not ~(switch)$}{
            \tcc{Craft a new sample}
            $\boldsymbol{x}' \leftarrow \boldsymbol{\Tilde{x}}^{(k)}$\;
            \For{$t=1, 2, \cdots, m$}{
                Uniformly select a set $\{c, w, h\}$ at random without replacement\;
                $\boldsymbol{x}'_{\text{B}_\text{t}} \leftarrow \boldsymbol{x}'_{[c,w-n:w+n,h-n:h+n]}$\;
                $\boldsymbol{x}_{\text{B}_\text{t}} \leftarrow \boldsymbol{x}_{[c,w-n:w+n,h-n:h+n]}$\;
               
                 \tcc{Perturbation region}
                $M \leftarrow \text{sign}(\boldsymbol{x}_{\text{B}_\text{t}} - \boldsymbol{x}'_{\text{B}_\text{t}})$\;
             $\boldsymbol{x}'_{\text{B}_\text{t}}\leftarrow \boldsymbol{x}'_{\text{B}_\text{t}} + M \times \delta$\;
            }
            
            \tcc{Evaluate crafted sample}
            \uIf{$D_{n_\text{q}} > D(\boldsymbol{x},\boldsymbol{x}')$}{
                $n_\text{q} \leftarrow n_\text{q}+1$\;
                \uIf{$\mathcal{C}(f(\boldsymbol{x}'))=1$}{
                    $\boldsymbol{\Tilde{x}}^{(k+1)} \leftarrow \boldsymbol{x}'$\;
                    $k \leftarrow k+1$ \;
                }
                $D_{n_\text{q}}\leftarrow D(\boldsymbol{x},\boldsymbol{\Tilde{x}}^{(k)})$\;
                Compute $\Delta$ using Equation \ref{eq:3}\;
                \uIf{$\Delta < \epsilon_\text{s}$}{$switch \leftarrow True$\;}
            }
            
            $j \leftarrow j + m$\; 
        }
        $\delta \leftarrow \frac{\delta}{\lambda}$\;
    }
    \KwRet{$\boldsymbol{\Tilde{x}}^{(k)}$}
    \caption{\RSBCD}
    \label{algo2}
\end{algorithm}

\subsection{\RSBCD}\label{sec:rsbcd} 

We recognize that the architecture of most machine learning models in computer vision is based on a Convolutional Neural Network (CNN) built on convolution operations. These convolution operations are defined as $c\times q \times q$ where $q$ is the size of the filter and $c$ is the number of channels to extract local patterns of an image. Consequently, we hypothesize that altering a block of coordinates as a square-shaped region with an appropriate size can target significant filter outputs potentially having a significant impact on the network's decision. Perturbing these coordinates can result in an adversarial example with fewer queries since we target regions of the input to impact actual convolutional filters and potentially discover salient features to mimic. 
Inspired by this, we adopt a notion of square-block perturbation regions and introduce \RSBCD that manipulates blocks of size $n$. \RSBCD has two stages: i) crafting a sample; and ii) its evaluation as described in Algorithm \ref{algo2}. 
\vspace{1mm}
\noindent\textbf{Crafting a Sample.~} In each iteration, the first stage of \RSBCD aims to yield a sample $\boldsymbol{x}'$ that is initialized with $\boldsymbol{x}^{(k)}$ which is an adversarial example at $k$-th step. To increase convergent rate and reduce query number, \RSBCD modifies several blocks of coordinates concurrently. It firstly selects $m$ different coordinates across different channels (R, G, B) of an image by choosing a set $S = \{S_1, S_2, \cdots, S_\text{m}\}$ where $S_\text{t} = \{c_\text{t}, w_\text{t}, h_\text{t}\}$ is selected uniformly at random such that $c_\text{t} \in [1, C]$, $w_\text{t} \in [1, W]$ and $h_\text{t} \in [1, H]$, where $t = 1, 2, \cdots, m$ and $C, W, H$ denote the number of channel, width and height of an image. 
This random selection is sampling without replacement and each selected coordinate $x'_\text{c,w,h}$ is a center of a square block $\boldsymbol{x}'_{\text{B}_\text{t}}$, where $\boldsymbol{x}'_{\text{B}_\text{t}}$ represents $\boldsymbol{x}'_{[c_\text{t},w_\text{t}-n:w_\text{t}+n,h_\text{t}-n:h_\text{t}+n]}$. 
Likewise, $m$ corresponding blocks $\boldsymbol{x}_{\text{B}_\text{t}}$ are yielded from the source image $\boldsymbol{x}$. 
A mask $M$ with the same size as  $\boldsymbol{x}'_{\text{B}_\text{t}}$ can be defined as $M=sign(\boldsymbol{x}_{\text{B}_\text{t}} - \boldsymbol{x}'_{\text{B}_\text{t}})$. This mask is used to identify the direction of perturbation for each element of a block $\boldsymbol{x}'_{\text{B}_\text{t}}$. When each element of a block which is a coordinate of an image is manipulated to move along this direction, it tends to move towards to its corresponding element in the source image. The sample $\boldsymbol{x}'$ is crafted when each of $m$ blocks of coordinates is updated as below:
\begin{equation}
    \label{eq:2} 
    \boldsymbol{x}'_{\text{B}_\text{t}}\leftarrow \boldsymbol{x}'_{\text{B}_\text{t}} + M \times \delta
\end{equation}

Where $\delta$ is a scalar which denotes an amount of perturbation for each element and it reduces by $\lambda$ after each cycle. One cycle is ended when all coordinates are selected for perturbation. If $\delta$ is initialized with a small value, it is slow convergent and results in query inefficiency from the beginning. Whilst, for large initial $\delta$, modifying blocks of coordinate almost leads to a sample moving further from the source image from beginning rather than moving closer. Consequently, it requires several cycles until $\delta$ reduces to a suitable value. To tackle this issue, we exploit the distribution of the absolute difference between all coordinates of a sample and their corresponding coordinate in a source image and use $i$-th percentile $P_\text{i}$ of this distribution to specify a proper initial $\delta$. In Equation \ref{eq:2}, only selected square blocks are perturbed while the rest of $\boldsymbol{\Tilde{x}}$ remains unchanged. 

\vspace{1mm}
\noindent\textbf{Evaluate Crafted Sample.~} In the second stage, to ensure a descent of distortion and improve query efficiency, a crafted sample $\boldsymbol{x}'$ is merely evaluated by the victim model if it moves closer to $\boldsymbol{x}$. If the adversarial criteria is then satisfied ($\mathcal{C}(f(\boldsymbol{x}'))=1$), the perturbation will make a change to update the next adversarial example as $\boldsymbol{\Tilde{x}}^{(k+1)} = \boldsymbol{x}'$. Otherwise the perturbation will be discarded.

\vspace{1mm}
\noindent\textbf{Determining when to switch to the next component.~} Similar to the switching criterion of gradient estimation methods, \RSBCD should switch to the next component when it cannot find any better adversarial example that can be empirically measured by distortion reduction rate $\Delta$ per $T$ queries. However, we observe that \RSBCD is a gradient-free optimization so $\Delta$ is highly varied for each subsequent query. 
As such we cannot simply apply the same criterion as gradient estimation methods. Consequently, to determine a better switching criterion for \RSBCD, we adopt a smoothing technique based on Simple Moving Average to measure the distortion reduction rate $\Delta$. In practice, $\Delta$ is computed as follows: 
\begin{equation}\label{eq:3}
    \Delta \leftarrow \frac{1}{T}\sum^{n_q-T}_{l=n_q-2T}(D_l - D_{(l+T)})
\end{equation}
where $D_\text{l}$ is a distance between $\boldsymbol{x}$ and $\boldsymbol{\Tilde{x}}^{(k)}$ at query $l$, $n_\text{q}$ is $n_\text{q}$-th query. If $\Delta$ is smaller than a switching threshold $\epsilon_\text{s}$, \RSBCD switch to the next component.

\section{Experiments and Evaluations} \label{sec:evaluations}
\subsection{Experiment Settings and Summary} \label{experiment setting}
\noindent\textbf{Attacks and Datasets.~} In this section, we evaluate the effectiveness of our \rambo versus current state-of-the-art attacks---Boundary attack (Boundary)~\cite{Brendel2018}, Sign-OPT \cite{Cheng2020} and HopSkipJump \cite{Chen2020} on two standard datasets: \texttt{CIFAR10}~\cite{Krizhevsky} and \texttt{ImageNet} \cite{Deng2009}. All hyper-parameters of our \rambo are described in Appendix \ref{apd-hyper-parameters} and all of the evaluation sets are described in Section~\ref{sec:Robustness of Hybrid Methods}, \ref{sec:benchmark on hard and non hard set}, Appendix \ref{apd-Robustness Evaluation Protocol} and \ref{apd-Validation on balance datasets}. 

\noindent\textbf{Models.~}\label{model description} For a fair comparison, for \texttt{CIFAR10}, we use the same CNN architecture used by Cheng et al. ~\cite{Cheng2019b, Cheng2020}. This network comprises of four convolutional layers, two max-pooling layers and two fully connected layers. 
For evaluation on \texttt{ImageNet}, we use a pre-trained ResNet-50 \cite{He2016} provided by torchvision \cite{Marcel2010} which obtains 76.15\% Top-1 test accuracy. In addition, all images are normalized into pixel scale of $[0,1]$.

\vspace{1mm}
\noindent\textbf{Evaluation Measures.~} To evaluate the performance of method, prior works use different metrics such as a score based on the median squared $l_2$-norm \cite{Brendel2018} and median $l_2$-norm distortion versus the number of queries ~\cite{Cheng2020, Chen2020}. However, median metric is not able to highlight the existence of the so-called \hard cases and their impact on the performance of an attack so the evaluation may be less reliable. Therefore, in addition to median, we report average $l_2$-norm distortion. We also adopt Attack Success Rate (ASR) used in 
\cite{Chen2020} to measures the attack success of crafted adversarial samples, obtained with a given query budget, at various distortion limits.

\noindent\textbf{Gradient Estimation Selection for \rambo.~}\label{apd:gradientestimation}
We apply two state-of-the-art gradient estimation methods, HopSkipJump and Sign-OPT, and derive two \rambo attacks: i)~\rambo(HSJA), composed of HopSkipJump, \RSBCD and Sign-OPT; and ii)~\rambo(SOPT),  composed of Sign-OPT and \RSBCD. We do not use HopSkipJump for the second gradient descent stage because we observe Sign-OPT to be more effective at refining adversarial samples---as also observed in~\cite{Cheng2020}. 

\noindent\textbf{Experimental Regime.~} We summarize the extensive experiments conducted with \texttt{CIFA10} and \texttt{ImageNet} datasets. All experiments are performed on one RTX TITAN GPU and one 2080Ti GPU. The total running time for all experiments is approximately 1,742 hours. The running time of each experiment is described in Appendix \ref{apd-experiment time summary}.

\begin{itemize}
    \item \textit{Robustness of \rambo:~}From the observations in Section~\ref{observations}, we aim to investigate the robustness of our \rambo by assessing the existence of a \hard set for our \rambo. We execute the exhaustive evaluation protocol used in Section~\ref{observations} and compare results with state-of-the-art attacks in Section \ref{sec:Robustness of Hybrid Methods}.
    
    \item \textit{Attacking \hardupper Sets:~}Most attacks demonstrate impressive performance in \easy cases whilst struggling with \hard cases. Therefore, 
    we compare and demonstrate the performance differences---in terms of query efficiency, attack success rate and distortion---that exists on \hard evaluation sets in Section~\ref{sec:benchmark on hard and non hard set}. 
   
    \item \textit{Impact of the Starting Image:~} We observed the impact of the starting image from the target class on the success of the attack in Section \ref{observations}. Hence, the exhaustive experimental evaluations in Section~\ref{sec:sensitivity-to-starting-image} explores the sensitivity of an attack's success to the choice of the attacker's starting image. An important consideration to evade detection through trial-and-error testing of starting images to find easy samples or when access to samples (source or target class) are restricted.
    
    \item \textit{Attack Insights:~} We observed clear correlations between perturbations yielded by our \rambo and salient regions of target images embedded inconspicuously in adversarial examples. We investigate these artifacts resulting from the localized perturbation method in BlockDescent in Section \ref{sec:visual explanations}. 
    
    \item \textit{Attacks Against Defended Models:~}Decision-based attacks are able to fool standard models. This naturally leads to the critical question of whether or not such attacks are able to bypass defensed models. Thus, the experiments in Section~\ref{sec:Attack against a Defended Models} aim to investigate the robustness of decision-based attacks against defense mechanisms.
    
    \item \textit{Validation on Balance Datasets:~}Constructing \hard and \easy sets for all decision-based attack methods through exhaustive evaluations to assess robustness is extremely time consuming. Therefore, we propose a reliable and reproducible attack evaluation strategy to validate robustness. We differ the proposed evaluation protocol and  results to Appendix \ref{apd-Validation on balance datasets} and release all of the constructed sets for comparisons in future studies.

    \item \textit{Untargeted Attack Validation:~}In addition to targeted attacks, for completeness, we evaluate our \rambo and other state-of-the-art attacks on \texttt{CIFAR10} and \texttt{ImageNet} under the untargeted attack setting. We defer these results to Appendix \ref{apd-untargeted attack}.
    
\end{itemize}

\begin{figure}[htb]
    \begin{center}
        \includegraphics[scale=0.48]{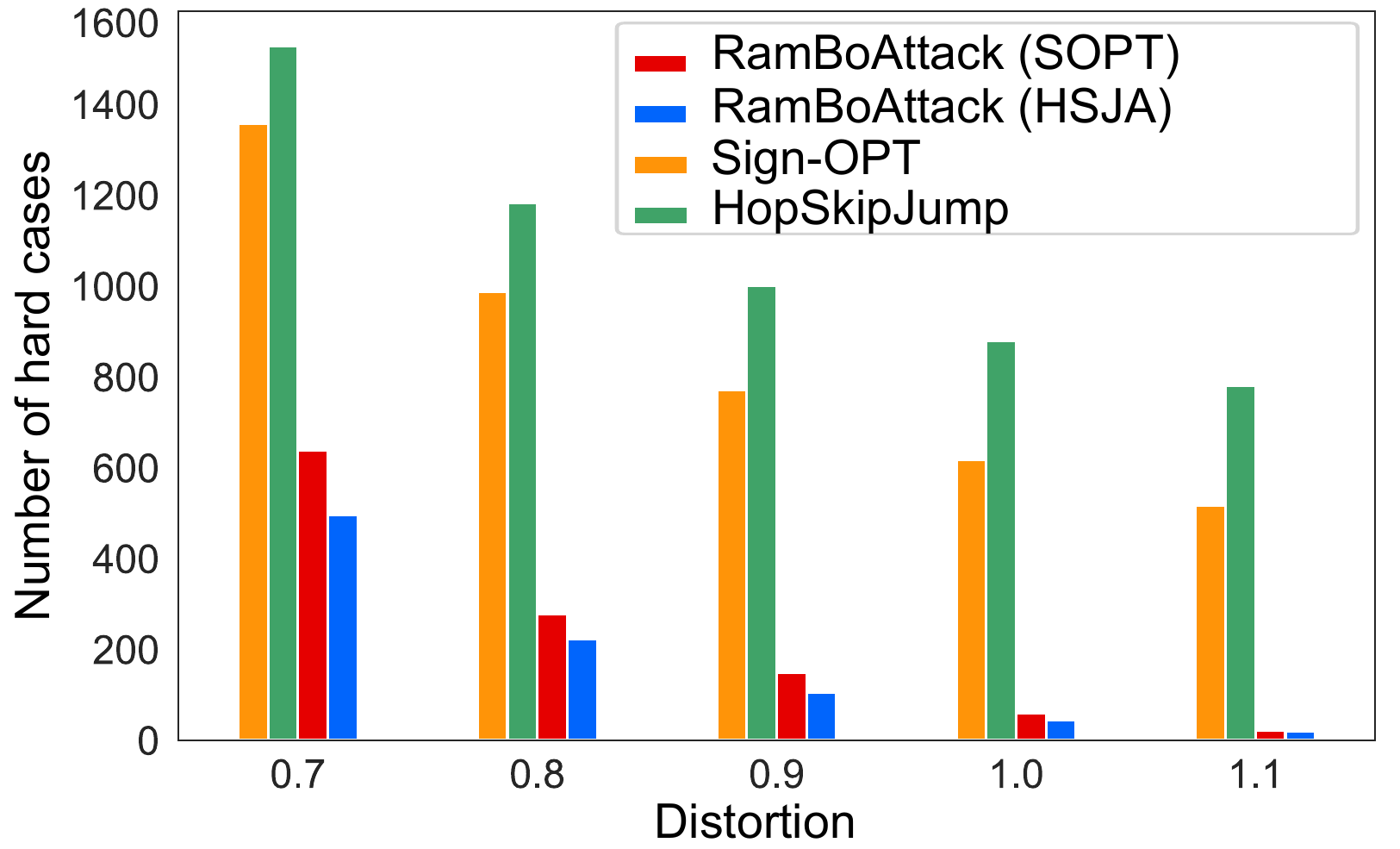}
        \caption{The number of \hard cases found for Sign-OPT, HopSkipJump and \rambo with a range of distortion threshold from 0.7 to 1.1 using a budget of 50,000 queries (see detailed results in Appendix~\ref{apd-Robustness of Hybrid Methods} and Fig.~\ref{fig:adp-Robustness of Hybrid Methods}).}
        \label{fig:barchart hard-case search}
    \end{center}
    \vspace{-3mm}
\end{figure}

\subsection{Robustness of \rambo} \label{sec:Robustness of Hybrid Methods}

We carry out a comprehensive experiment, similar to that in Section \ref{observations}. In this experiment we use a range of distortion threshold of 0.7 to 1.1. Notably, both \cite{Chen2020} and \cite{Cheng2020} reported their methods to achieve a distortion level below 0.3 after 10,000 queries; hence our proposed values are not guaranteed to discover \hard cases because the smallest value, 0.7, is much higher than 0.3 achieved in other studies. The main aim is to illustrate how our \rambo are able to craft more adversarial example with distortions below a range of distortions from 0.7 to 1.1 for each sample of the entire \texttt{CIFAR10} test set. We compare the performance of the \rambo with Sign-OPT and HopSkipJump. Fig. \ref{fig:barchart hard-case search} shows a remarkably low number of \hard cases for the \rambo. The total number of hard cases achieved for our \rambo is approximately 10 times lower for the distortion ranges from 0.9 to 1.1. For distortion at 0.7 and 0.8, the number of \hard cases drops approximately 2 times and 5 times, respectively in comparison with the other attack methods. Interestingly, as expected,  \hard pairs encountered by Sign-OPT and HopSkipJump are resolved with \rambo as shown in Appendix~\ref{apd-Robustness of Hybrid Methods}---see Fig.~\ref{fig:adp-Robustness of Hybrid Methods}.

\subsection{Attacking \hardupper Sets} \label{sec:benchmark on hard and non hard set} 

\begin{figure}[htb]
    \begin{center}
        \includegraphics[scale=0.46]{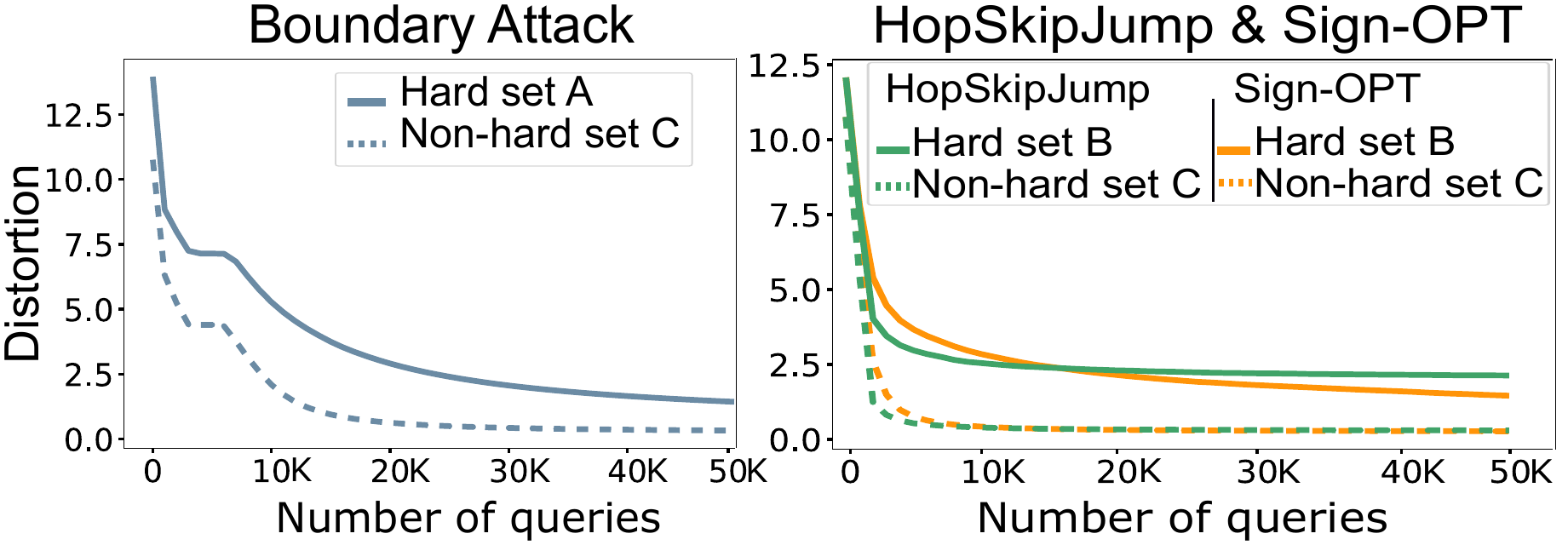}
        \caption{A distortion comparison versus queries for each method using their own \hard versus \easy cases.}
        \label{fig:hard-set result}
    \end{center}
    \vspace{-3mm}
\end{figure}

\begin{figure*}[htb]
    \begin{center}
        \includegraphics[scale=0.3] {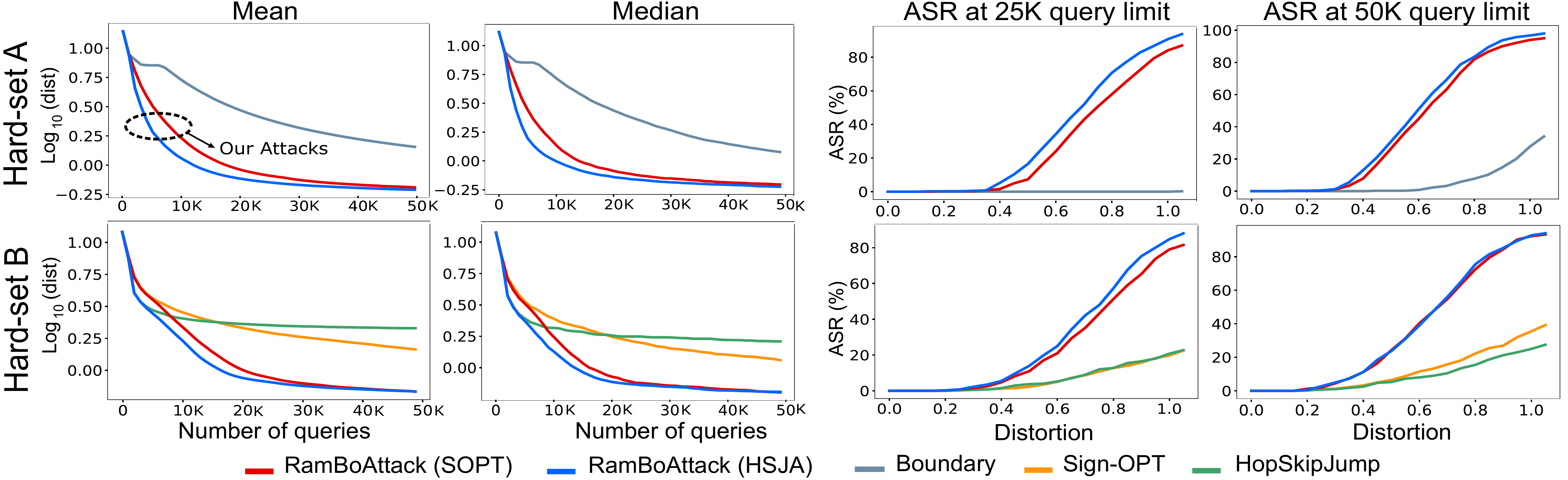}
        \caption{Distortion (dist) on a $\log_{10}$ scale vs number of queries. The first row shows the results for our \rambos versus Boundary attack on \hardset~A whilst the second row illustrates the results for our \rambos versus HopSkipJump and Sign-OPT on \hardset~B. Our \rambos are \textbf{\textit{more query efficient}} in \hard cases. Hence our attack is demonstrably more robust and query efficient.}
        \label{fig:benchmark-CIFAR10}
    \end{center}
    \vspace{-5mm}
\end{figure*}
\noindent\textbf{Evaluations on \texttt{CIFAR10}.~}From \texttt{CIFAR10} test set, we generate a \hard set for Boundary Attack called \hardset A and another \hard set for both Sign-OPT and HopSkipJump called \hardset B. The \hardset A and B are composed of 400 \hard sample pairs of a source image and a starting image. A \hard sample is selected when a distortion between a source image and its adversarial example found after 50,000 queries is larger than or equal to 0.9. For a fair comparison, each method is employed to craft an adversarial example for each source image initialized with a given starting image. In addition, we also construct a common \easy set for all three attacks called \easyset C to compare and highlight the significant difference between evaluation results on \hard and \easy sets as shown in Fig.~\ref{fig:hard-set result}. In particular, Fig.~\ref{fig:hard-set result} illustrates that the average distortion versus queries on the common \easyset~C achieved by these methods is significantly lower than that obtained on theirs own \hard set after 50,000 queries.  

\begin{figure}[htp]
    \begin{center}
        \includegraphics[scale=0.32]{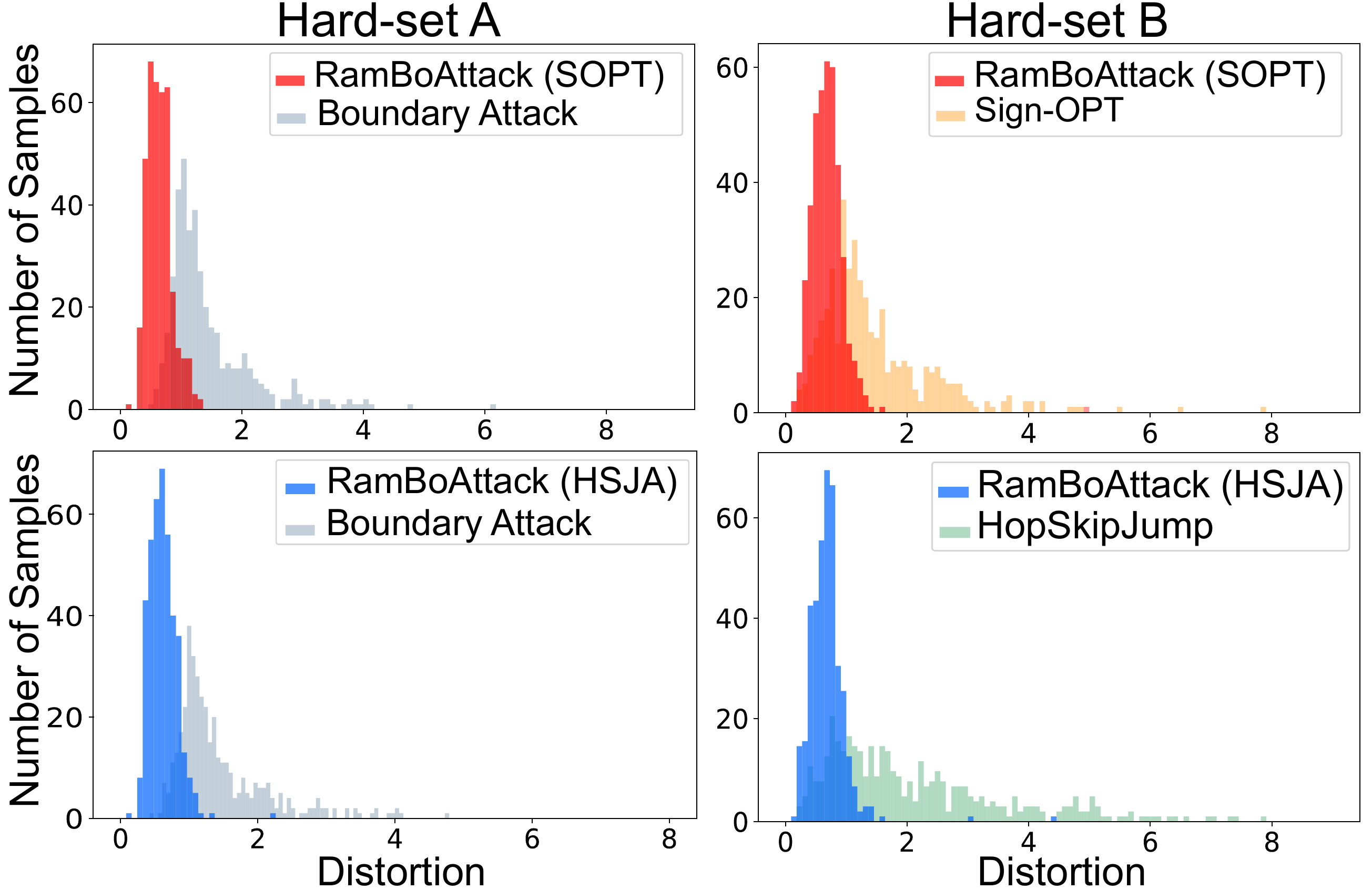}
        \caption{On both \hardset A and B selected from \texttt{CIFAR10}, the distortion distribution yielded by our \rambos are shifted left and indicates an overall  smaller distortion compared to other attacks.}
        \label{fig:hard set hist}
    \end{center}
    \vspace{-3mm}
\end{figure}

We evaluate our \rambo on \hardset A \& B. Fig.~\ref{fig:benchmark-CIFAR10} shows that Boundary Attack, Sign-OPT and HopSkipJump do not efficiently find an adversarial example with low distortion; however, \rambo can achieve better performance on the \hardsets. We defer detailed evaluations on \easy-sets to Appendix~\ref{apd-Validation on balance datasets}; as expected, \rambo performs comparably-well on these sets. Histogram charts in Fig.~\ref{fig:hard set hist} demonstrate that for each \hardset, our attacks are able to find lower distortion adversarial examples for most \hard cases and the distortion distribution on both \hardsets: i)~are shifted to smaller distortion regions; and ii)~show significantly smaller spread or variance. 

\begin{figure}[htb]
    \begin{center}
        \includegraphics[scale=0.36]{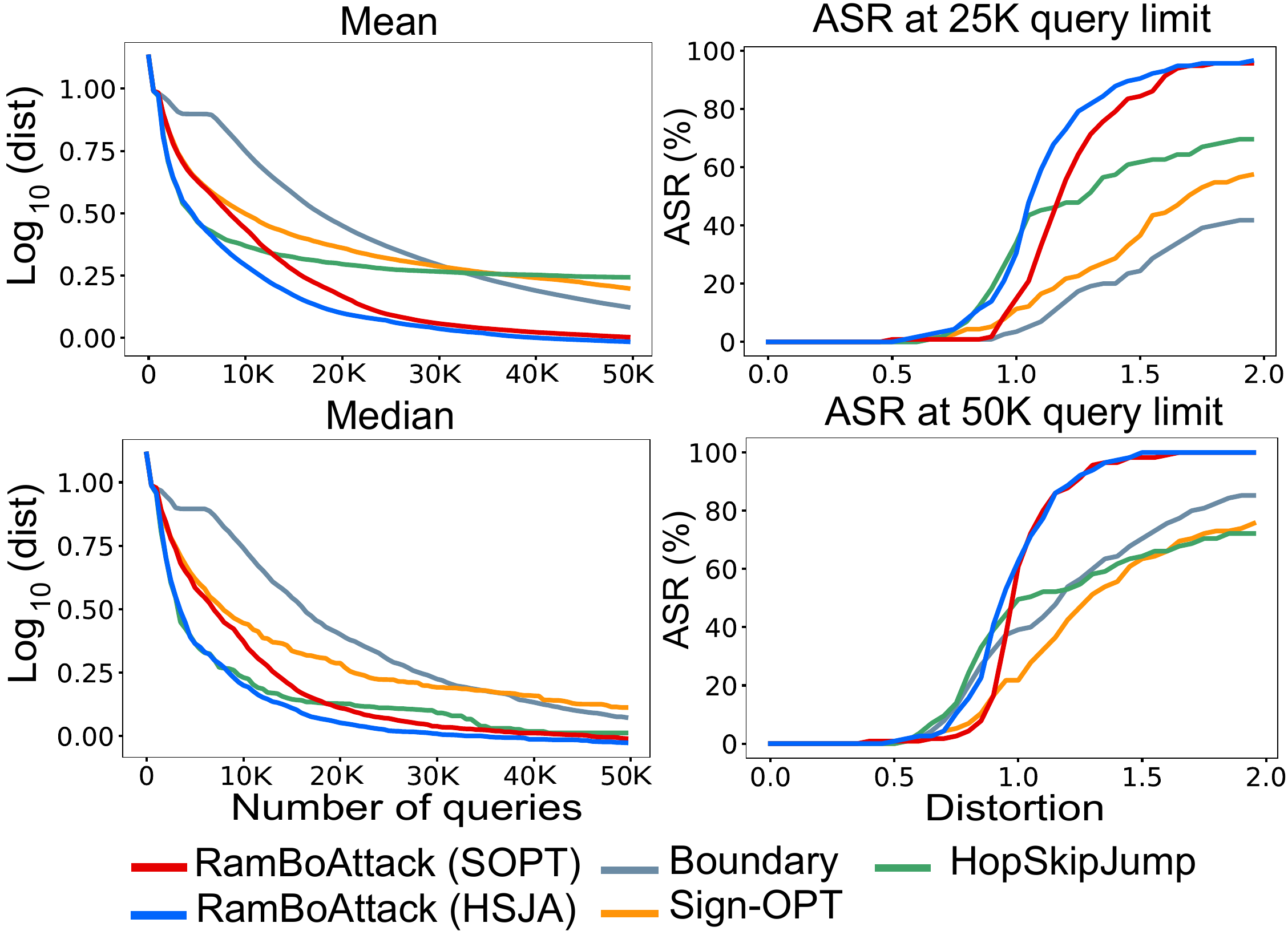}
        \caption{Distortion in a $\log_{10}$ scale vs number of queries on \hardset-D. 
        Our \rambo is \textbf{\textit{more query efficient}} and achieves a higher ASR on this \hardset. Hence, our attack is demonstrably more robust and query efficient.}
        \label{fig:hardset - Rambo}
    \end{center}
    \vspace{-3mm}
\end{figure}

Although we observe \rambo to result in fewer hard samples in comparison to other methods at various distortion thresholds, we construct a \hard set for \rambo called \hardset D based on the same criteria used to generate \hardset A and B to assess if the \hardset for \rambo could somehow be easier for the other attack methods. The total number of samples for this set is 115 sample pairs because \rambo has a much lower number of \hard cases than their counterparts (namely BA, HopSkipJump and Sign-OPT) at a given distortion threshold as illustrated in Fig.~\ref{fig:barchart hard-case search}. We summarize the results from our evaluations in Fig.~\ref{fig:hardset - Rambo}. As expected, \rambos are more query efficient and are able to craft lower mean and median distortion adversarial examples as well as achieve higher attack success rates at both query budgets. In particular, at distortion levels above 1.0, in comparison to other attacks, \rambos obtain much higher attack success rates---notably, with significant margins at the lower query budget of 25K, since \rambos employ \RSBCD when the gradient estimation method is unable to make progress (potentially being stuck in a bad local minimum), to discover better solutions and craft lower distortion adversarial samples. 

\begin{figure}[htb]
    \begin{center}
        \includegraphics[scale=0.3]{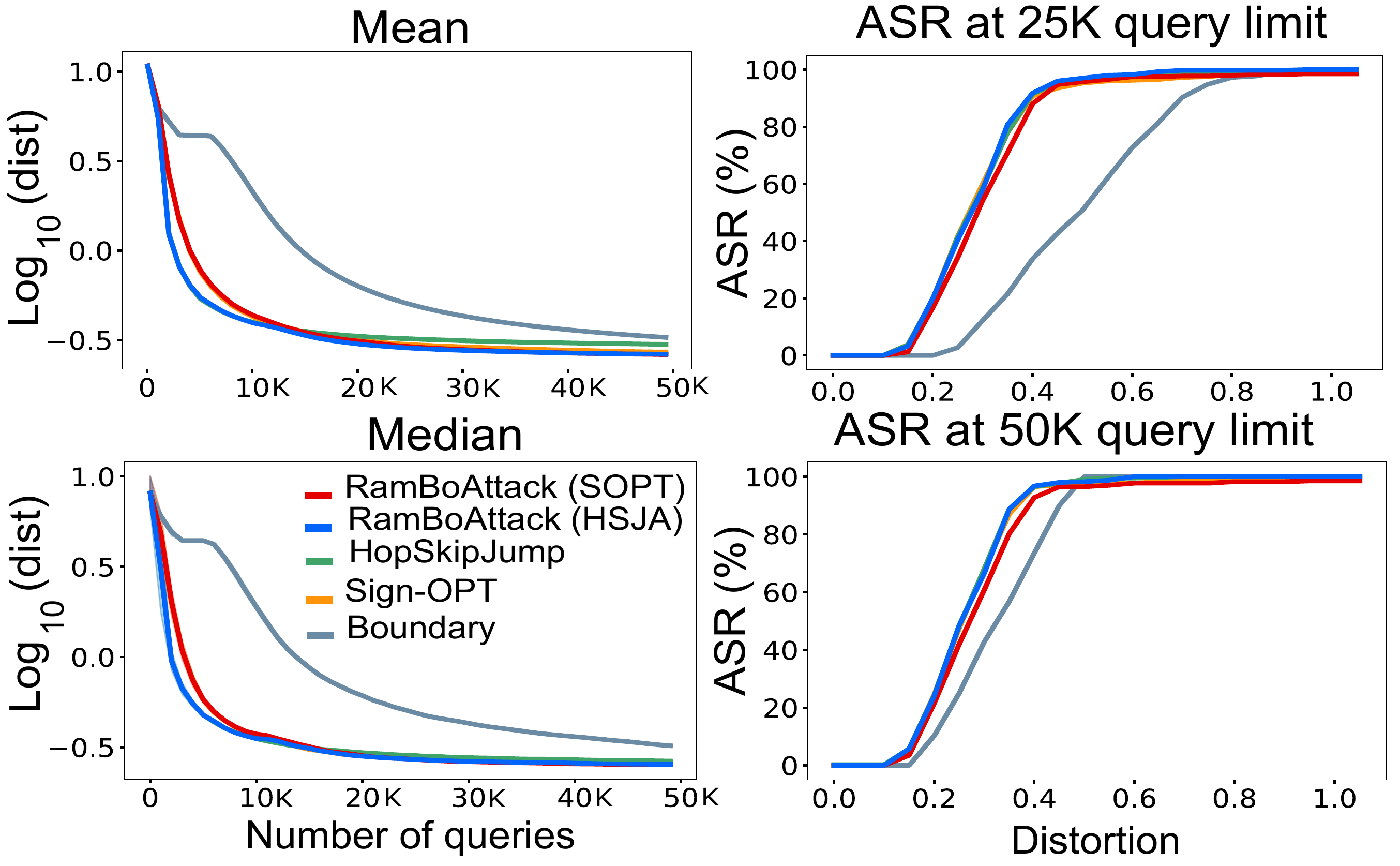}
        \caption{Distortion (dist) in a $\log_{10}$ scale vs number of queries on \hard \texttt{ImageNet} evaluation sets. It shows the results on the \hardset and our \rambos are \textbf{\textit{more query efficient}}. Hence our attack is demonstrably more robust and query efficient.}
        \label{fig:benchmark-ImgN}
    \end{center}
    \vspace{-3mm}
\end{figure}

\begin{figure}[ht]
    \begin{center}
        \includegraphics[scale=0.32]{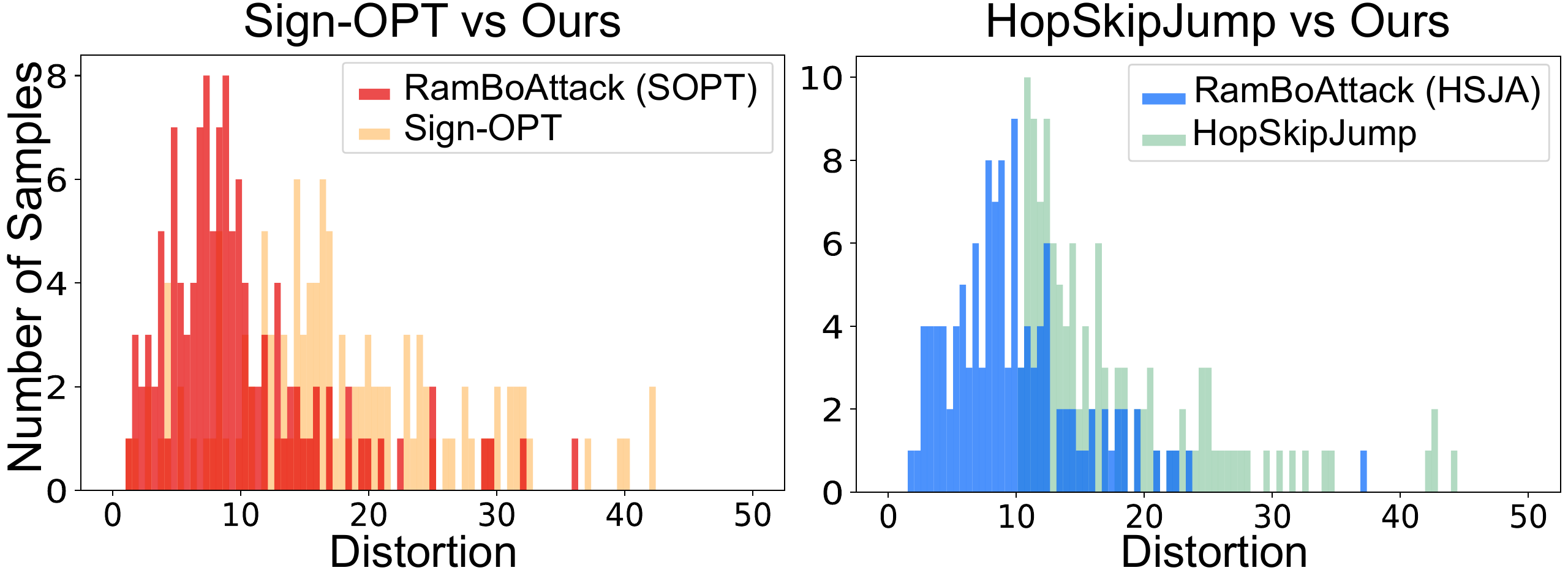}
        \caption{On the \hardset selected from \texttt{ImageNet}, the distortion distributions yielded by our \rambos indicate an overall smaller distortion compared to other attacks. The distributions is shifted to the left and has significantly less variance compared to other attacks.}
        \label{fig:ImgN Hist}
    \end{center}
    \vspace{-3mm}
\end{figure}

\noindent\textbf{Evaluation on \texttt{ImageNet}.} To demonstrate the robustness of our attacks on a large scale model and dataset, we compose a \hardset with $120$ \hard sample pairs from \texttt{ImageNet}. A hard sample  is  selected  when  a  distortion  between  a  source image and its adversarial example found after 50,000 queries by Sign-OPT and HopSkipJump is  larger than or equal to $15$. Notably, we do not compose a \hard set for Boundary Attack because it cannot yield low distortion adversarial examples efficiently on large scale datasets. Fig. \ref{fig:benchmark-ImgN} demonstrates that our \rambos outperform both Sign-OPT and HopSkipJump on the \hardset. We defer detailed evaluations on \easy-\textit{sets} to Appendix~\ref{apd-Validation on balance datasets}; notably, \rambos achieve improved results on the more complex \texttt{ImageNet} dataset. The histograms in Fig.~\ref{fig:ImgN Hist} show distortion distributions for our attacks shifted significantly to smaller distortion regions with smaller variance and fewer outliers compared to other attacks.

\subsection{Impact of Various Starting Images} \label{sec:sensitivity-to-starting-image}

\begin{figure}[htb]
    \begin{center}
        \includegraphics[scale=0.7]{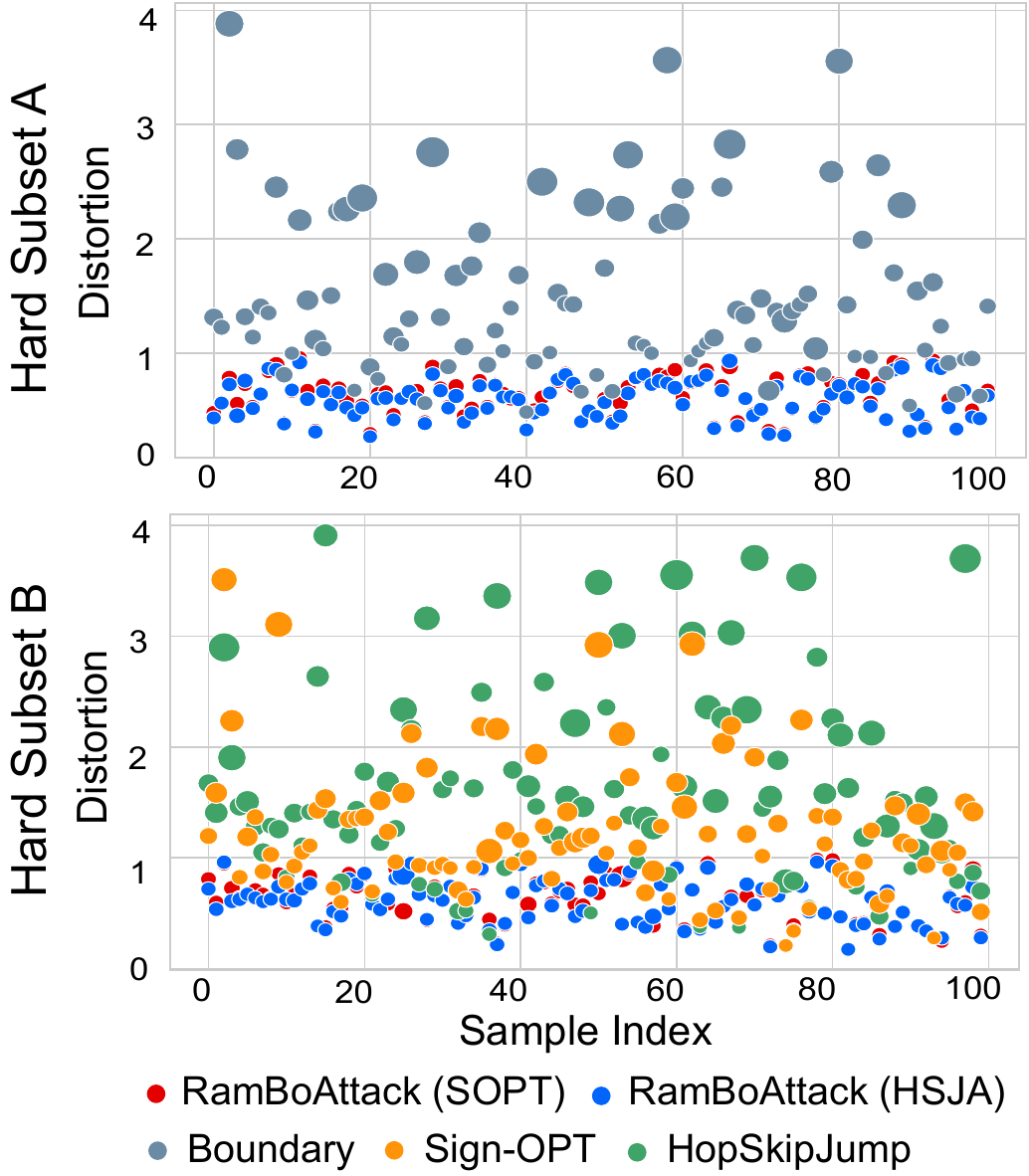}
        \caption{An illustration of sensitivity of different attacks to various starting images. Each method is evaluated on each subset and the charts show the average and variance of distortion for each case of each subset achieved by different methods. $y$-axis denotes the average distortion while the size of each bubble denotes the distortion variation. Compared with Boundary, Sign-OPT and HopSkipJump attacks, our \rambos are \textbf{\textit{much less sensitive to the choice of a starting image}}.}
        \label{fig:one vs various starting image-hard-easy set}
    \end{center}
    \vspace{-3mm}
\end{figure}
In this experiment, we first compose \textit{subset} A and B by selecting $100$ random \hard sample pairs from \hardset A and B, respectively (see Section \ref{sec:benchmark on hard and non hard set} for these sets). Our \rambos are compared with Boundary attack on subset A and with Sign-OPT and HopSkipJump, on subset B. In Section \ref{sec:benchmark on hard and non hard set}, each method needs to yield an adversarial example for a pair of a given source image and a given starting image. In contrast, in this experiment, the given starting image is replaced by 10 starting images randomly selected from the \texttt{CIFAR10} evaluation set and correctly classified by the model. All evaluations are executed with a 50,000 query budget.

In Fig.~\ref{fig:one vs various starting image-hard-easy set}, the size of each bubble denotes the standard deviation while y-axis value indicates average distortion. We can see that our \rambos almost achieve smaller mean and standard deviation than Sign-OPT, HopSkipJump and Boundary Attack on subset A and B. A robust method should be less susceptible to the selection of a starting image and yield a low distortion adversarial example most chosen starting images. We can observe from Fig.~\ref{fig:one vs various starting image-hard-easy set} that our \rambos are more robust than Sign-OPT, HopSkipJump and Boundary attacks as a consequence of being less sensitive to the chosen starting images. We also carry out this experiment on an \easy subset C and defer these results to Appendix~\ref{apd:Impact of Starting Images}.

\subsection{Attack Insights} \label{sec:visual explanations}

\noindent\textbf{Perturbation Regions.~}
First, we develop a simple technique to transform a perturbation with size $C \times W \times H$ to a Perturbation Heat Map (PHM) with size $W \times H$ that is able to visualize perturbation magnitude of each pixel. This transformation is defined as:

\begin{equation}\label{eq:4}
\begin{aligned}
PHM_{i,j} \leftarrow  \frac{\text{A}_{i,j}}{\displaystyle\max(\text{A})},
\end{aligned}
\end{equation} 

where $\text{A}_{i,j} = \sum^{C}_{c=1}|(\boldsymbol{x}-\boldsymbol{x_\text{a}})_{c,i,j}|$; $c \in  [1, C]$, $i \in [1, W]$ and $j \in [1, H]$. Second, since Grad-CAM \cite{Selvaraju2017} is a popular visual explanation technique for visualizing salient features in an input image to understand a CNN model's decision, we use it to investigate the adversarial perturbations generated by our attack and the salient features in the target image largely responsible for a model's decision for the classification of an input to a target class. 

\begin{figure}[htp]
    \begin{center}
\includegraphics[scale=0.36]{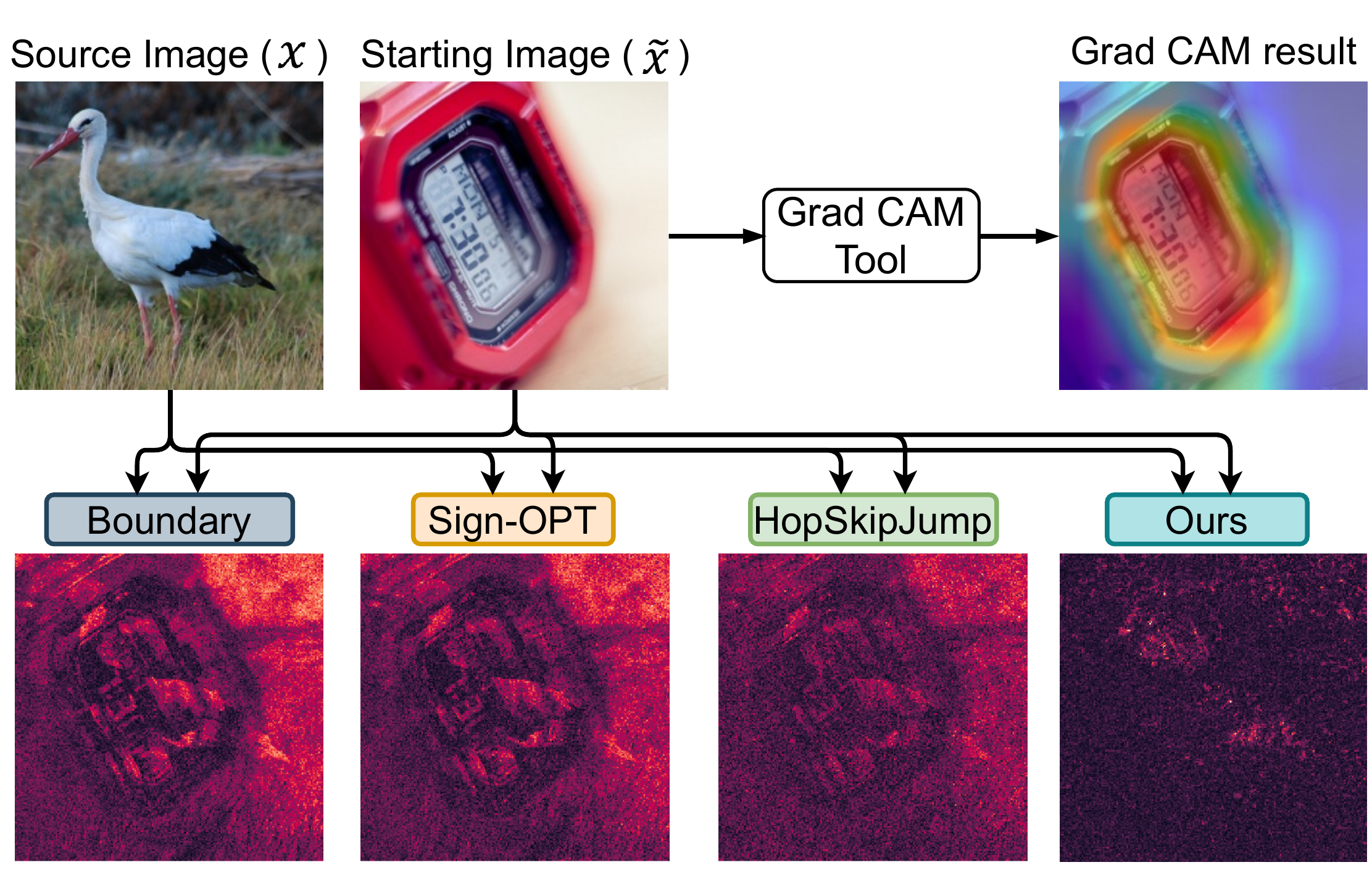}
        \caption{Grad-CAM tool visualizes salient features of the starting image or target class:~\texttt{digital watch}. Perturbation heat map (PHM) visualizes the normalized perturbation magnitude at each pixel.  Comparing different perturbations crafted by different attacks highlights that the localized perturbations yielded by \rambo concentrate on salient areas illustrated by GRAD-CAM and embeds these targeted perturbations in the source image to fool the classifier to predict the target class; even though, \rambo does not exploit the knowledge of salient regions to generate perturbations---additional examples in Appendix Fig.~\ref{fig:apd-GradCAM-visual}}
        \label{fig:GradCAM-visual}
    \end{center}
    \vspace{-3mm}
\end{figure}

In all of the attack methods, we observe the attacks to embed the target image in the source image in a deceptive manner. However, in \hard cases, based on PHM and Grad-CAM outcomes, we observe a strong connection between  adversarial perturbations found and salient regions in starting images as illustrated in Fig. \ref{fig:GradCAM-visual} for \rambos. It shows that our \rambos are able to discover and limit manipulations of pixels to salient regions responsible for determining the classification decision of an input image to the target class to craft adversarial examples. This salient region consists of the most discriminative local structures of a starting image against a source image. Because BlockDescent is able to manipulate local regions, \rambos are able to  exploit only this discriminative region and employ less adversarial perturbations than Sign-OPT and HopSkipJump to promote features of a starting image and suppress the feature of the source image. Therefore, it may shed light on why \rambo with the core component \RSBCD are able to tackle the so-called \hard cases. Moreover, in these \hard cases, we observe that our \rambo is able to yield perturbations with more semantic structure if compared with Sign-OPT or HopSkipJump.

\begin{figure*}[!t]
    \begin{center}
        \includegraphics[scale=0.2]{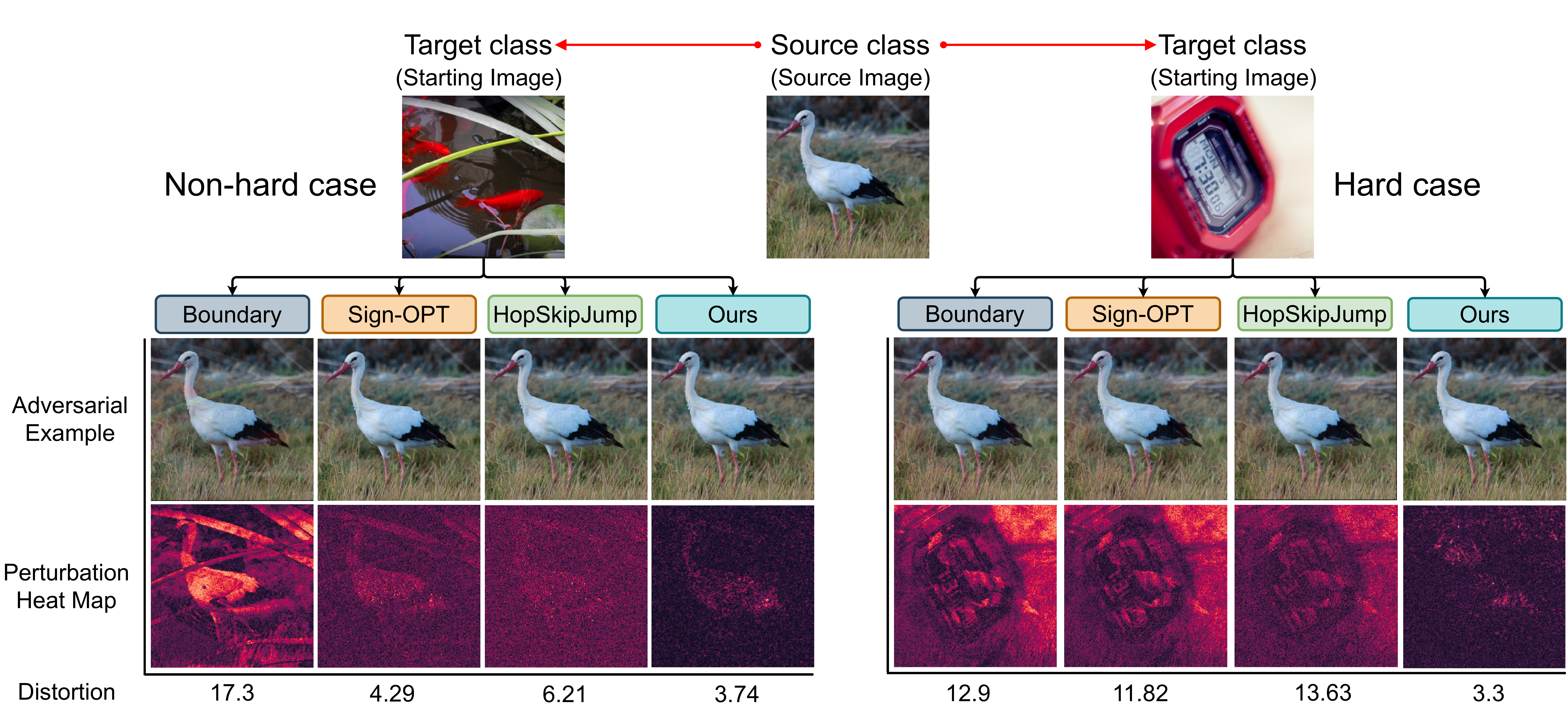}
        \caption{An illustration of \hard case (\texttt{white stork} to \texttt{goldfish}) versus \easy case (\texttt{white stork} to \texttt{digital watch}) on \texttt{ImageNet}. Adversarial examples in \easy cases and \hard cases are yielded after 50K and 100K queries, respectively. Except for Boundary attack, adversarial examples crafted by different attacks in \easy cases are slightly different whilst in the \hard case, our \rambo is able to craft an adversarial example with much smaller distortion than other attacks due to the ability of our BlockDescent formulation to \textit{target effective localized perturbations}.}
        \label{fig:ImageNet hard-easy}
    \end{center}
    \vspace{-5mm}
\end{figure*} 

\noindent\textbf{Visualization of \texttt{ImageNet} Hard versus Non-hard Cases.~}Fig. \ref{fig:ImageNet hard-easy} illustrates adversarial examples in \easy cases and \hard cases yielded by Boundary Attack, Sign-OPT, HopSkipJump and our \rambo(HSJA) after 50K and 100K queries, respectively. The second row of Fig. \ref{fig:ImageNet hard-easy} shows each corresponding adversarial example and the third row illustrates PHM of each adversarial example. The last row shows the $l_2$ distortion between each adversarial example and the source image.

For the adversarial example of \easy cases, all methods are able to craft low distortion adversarial examples except Boundary attack. These adversarial examples and their corresponding distortions are comparable. On the contrary, adversarial examples in \hard cases yielded by Boundary Attack, Sign-OPT and HopSkipJump have \textit{noticeably higher distortion} than the one crafted by our attack. We observe Boundary Attack, Sign-OPT and HopSkipJump to experience potential entrapment when searching for a low distortion adversarial example, even when the budget is increased to 100,000 queries. 

\vspace{1mm}
\noindent\textbf{Convergence.} \label{apd-Convergence analysis} The problem considered in this paper is non-convex and non-differentiable. As such, providing a guaranteed global minimum is not possible. However, our insight is that the gradient estimation in blackbox attacks is unreliable particularly in the vicinity of the local minima. To remedy the problem,  we propose \rambo as a generic method to overcome this issue. We employ a gradient estimation method in the initial descent using any of the existing alternatives (before \RSBCD) and subsequently in the refinement stage (after \RSBCD). 
Hence, employing the gradient estimation in \cite{Cheng2020}, for instance, would imply that the theoretical convergence analysis therein is still valid for our method.

\subsection{Attack Against Defense Mechanism} \label{sec:Attack against a Defended Models}
In this section, we evaluate the robustness of various attacks against three different defense mechanisms including region-based classification, adversarial training and defensive distillation. We choose these defense methods due to their own strength; to illustrate, region-based classifiers can pragmatically alleviate various adversarial attacks without sacrificing classification accuracy on benign inputs~\cite{Cao2017} whilst adversarial training ~\cite{Goodfellow2014, Madry2017,Tramer2018} is one of the most effective defense mechanisms against adversarial attacks~\cite{Athalye2018} and defensive distillation~\cite{Papernot2016a} employ's a form of gradient masking.  

For a \textit{baseline}, we choose C$\&$W attack \cite{Carlini2017}, a state-of-the-art \textit{white-box attack}. The adversarial training based models used in this experiment is trained with Projected Gradient Descent (PGD) adversarial training proposed in~\cite{Madry2017}. The experiment is conducted on the balance set withdrawn from \texttt{CIFAR10} described in Appendix \ref{apd-Validation on balance datasets}. We evaluate our \rambo and current state-of-the-art decision-based attacks at different query budgets: 5K, 10K, 25K and 50K.  

Based on the results, shown in Appendix~\ref{apd-defended model} for more details, we observe that \rambos are \textit{more robust} than Boudnary, Sign-OPT, HopSkipJump and even C$\&$W (\textit{white-box attack baseline}) when attacking a region-based classifier. In attacks against models using adversarial training and defensive distillation, \rambos are able to achieve comparable performance to Sign-OPT and HopSkipJump but outperform Boundary and C$\&$W attack---\textit{white-box} attack baseline.

\section{Related Work}
\noindent\textbf{Transfer Approaches.~} Malicious adversaries are able to exploit transferability of adversarial example generated on an ensemble DNN to attack against a target neural network as shown by Liu et al. \cite{Liu2017}. Papernot et al. \cite{Papernot2017a} introduced a transfer attack by training a surrogate model with output queried from a target model. Even though this approach does not require  prior knowledge and full access to a model, it must have access to a full or partial training dataset in order that they can train a surrogate model to synthesize adversarial examples. Moreover, for complex target models, the transfer approach has limited effectiveness \cite{Suya2020}. 

\vspace{1mm}
\noindent\textbf{Random Search Approaches.~}
In decision-based setting, Brendel et al. \cite{Brendel2018} and Brunner et al. \cite{Brunner2019} proposed Boundary Attack (BA) and Biased Boundary Attack (Biased BA) respectively that require limited information and access to a target DNN model such as top-k predicted labels. Instead of searching on Gaussian distribution like BA, Biased BA exploits low frequency perturbations based on Perlin Noise and combines with regional masking as well as gradients from surrogate models. Even though both of them work surprisingly well, they do not gain query efficiency and require a large number of queries to explore a high-dimensional space. Another attack method introduced by Ilyas et al. \cite{Ilyas2018} exploits discretized score based on the ranking of the adversarial label. However, since this method requires top k sorted label results from a deep learning model to estimate the discretized score, it cannot work in top 1 label scenario like BA or Bias BA. 

\vspace{1mm}
\noindent\textbf{Optimization Approaches.~}
In score-based scenario, attackers can query a deep learning model to receive probability outputs or confident scores. Therefore, Chen et al. \cite{Chen2017} can formulate an optimization problem to directly optimize an objective function based on these outputs. This method is considered as a derivative-free optimization method. Nevertheless, in the decision-based setting, adversaries have no access to confident scores or class probability to gain gradient information. Hence, the formulated optimization problem proposed by Chen et al. \cite{Chen2017} cannot be applied. In Section \ref{sec:SOTA approach} we discuss in detail optimization-based framework under decision-based setting and refer the reader to the section for further details.

\section{Conclusion}
Overall, we propose a new attack method in a decision based setting; \rambo. In contrast to modifying a whole image as in current attacks, we exploit localized perturbations to yield more effective and low distortion adversarial examples in the so-called \hard cases. Our empirical results demonstrate that our attack outperforms current state-of-the-art attacks. Interestingly, while the main proposed component, \RSBCD, is able to significantly improve the performance and robustness of attacks in the so-called \hard cases, it does not degrade performance in \easy cases. As a result, validation results on small and large scale evaluation sets demonstrate that \rambo is \textit{more robust and query efficient} than current state-of-the-art attacks. 

\bibliographystyle{IEEEtranS}
\bibliography{IEEEabrv,references}

\begin{thebibliography}{10}
\providecommand{\url}[1]{#1}
\csname url@samestyle\endcsname
\providecommand{\newblock}{\relax}
\providecommand{\bibinfo}[2]{#2}
\providecommand{\BIBentrySTDinterwordspacing}{\spaceskip=0pt\relax}
\providecommand{\BIBentryALTinterwordstretchfactor}{4}
\providecommand{\BIBentryALTinterwordspacing}{\spaceskip=\fontdimen2\font plus
\BIBentryALTinterwordstretchfactor\fontdimen3\font minus
  \fontdimen4\font\relax}
\providecommand{\BIBforeignlanguage}[2]{{%
\expandafter\ifx\csname l@#1\endcsname\relax
\typeout{** WARNING: IEEEtranS.bst: No hyphenation pattern has been}%
\typeout{** loaded for the language `#1'. Using the pattern for}%
\typeout{** the default language instead.}%
\else
\language=\csname l@#1\endcsname
\fi
#2}}
\providecommand{\BIBdecl}{\relax}
\BIBdecl

\bibitem{Amazon}
\BIBentryALTinterwordspacing
{"Amazon Machine Learning"}. [Online]. Available:
  \url{https://aws.amazon.com/machine-learning/}
\BIBentrySTDinterwordspacing

\bibitem{azureCS}
\BIBentryALTinterwordspacing
{"Azure Cognitive Service"}. [Online]. Available:
  \url{https://azure.microsoft.com/en-us/services/cognitive-services/}
\BIBentrySTDinterwordspacing

\bibitem{googlecloudvision}
\BIBentryALTinterwordspacing
{"Google Cloud Vision"}. [Online]. Available:
  \url{https://cloud.google.com/vision}
\BIBentrySTDinterwordspacing

\bibitem{IBMWatson}
\BIBentryALTinterwordspacing
{"IBM Watson Machine Learning"}. [Online]. Available:
  \url{https://www.ibm.com/cloud/machine-learning}
\BIBentrySTDinterwordspacing

\bibitem{Athalye2018}
A.~Athalye, N.~Carlini, and D.~Wagner, ``{Obfuscated gradients give a false
  sense of security: Circumventing defenses to adversarial examples},''
  \emph{International Conference on Machine Learning (ICML)}, 2018.

\bibitem{Brendel2018}
W.~Brendel, J.~Rauber, and Bethge, ``{Decision-based adversarial attacks:
  Reliable attacks against black-box machine learning models},''
  \emph{International Conference on Learning Recognition(ICLR)}, 2018.

\bibitem{Brunner2019}
T.~Brunner, F.~Diehl, M.~Le, and A.~Knoll, ``{Guessing smart: Biased sampling
  for efficient black-box adversarial attacks},'' \emph{The IEEE International
  Conference on Computer Vision (ICCV)}, 2019.

\bibitem{Cao2017}
X.~Cao and N.~Z. Gong, ``{Mitigating Evasion Attacks to Deep Neural Networks
  via Region-based Classification},'' \emph{Annual Computer Security
  Applications Conference (ACSAC)}, 2017.

\bibitem{Carlini2017}
N.~Carlini and D.~Wagner, ``{Towards evaluating the robustness of neural
  networks},'' \emph{IEEE Symposium on Security and Privacy}, 2017.

\bibitem{Chen2015}
C.~Chen, A.~Seff, A.~Kornhauser, and J.~Xiao, ``{DeepDriving: Learning
  Affordance for Direct Perception in Autonomous Driving},'' \emph{The IEEE
  International Conference on Computer Vision (ICCV)}, 2015.

\bibitem{Chen2020}
J.~Chen, M.~I. Jordan, and M.~J. Wainwright, ``{HopSkipJumpAttack: A
  Query-Efficient Decision-Based Attack},'' \emph{IEEE Symposium on Security
  and Privacy (SSP)}, 2020.

\bibitem{Chen2017}
P.~Chen, H.~Zhang, Y.~Sharma, J.~Yi, and C.~Hsieh, ``{Zoo: Zeroth order
  optimization based black-box attacks to deep neural networks without training
  substitute models},'' \emph{ACM Workshop on Artificial Intelligence and
  Security (AISec)}, pp. 15--26, 2017.

\bibitem{Cheng2019b}
M.~Cheng, T.~Le, P.~Chen, H.~Zhang, C.~Hsieh, and J.~Yi, ``{Query-Efficient
  Hard-label Black-box Attack: An Optimization-based Approach},''
  \emph{International Conference on Learning Recognition(ICLR)}, 2019.

\bibitem{Cheng2020}
M.~Cheng, S.~Singh, P.~Chen, P.-Y. Chen, H.~Yi, J.~Zhang, and C.-J. Hsieh,
  ``{Sign-OPT: A Query-Efficient Hard-label Adversarial Attack},''
  \emph{International Conference on Learning Recognition(ICLR)}, 2020.

\bibitem{Cheng2019}
S.~Cheng, Y.~Dong, T.~Pang, H.~Su, and J.~Zhu, ``{Improving Black-box
  Adversarial Attacks with a Transfer-based Prior},'' \emph{Conference on
  Neural Information Processing Systems (NeurIPS)}, 2019.

\bibitem{Deng2009}
J.~Deng, W.~Dong, R.~Socher, L.~Li, K.~Li, and L.~Fei-Fei, ``{ImageNet: A
  large-scale hierarchical image database},'' \emph{Computer Vision and Pattern
  Recognition(CVPR)}, 2009.

\bibitem{Goodfellow2014}
I.~J. Goodfellow, J.~Shlens, and J.~Szegedy, ``{Explaining and harnessing
  adversarial examples},'' \emph{International Conference on Learning
  Recognition(ICLR)}, 2014.

\bibitem{He2016}
K.~He, X.~Zhang, S.~Ren, and J.~Sun, ``{ Deep residual learning for image
  recognition},'' \emph{Computer Vision and Pattern Recognition (CVPR)}, p.
  770–778, 2016.

\bibitem{Ilyas2018}
A.~Ilyas, L.~Engstrom, A.~Athalye, and J.~Lin, ``{Black-box adversarial attacks
  with limited queries and information},'' \emph{International Conference on
  Machine Learning (ICML)}, 2018.

\bibitem{Krizhevsky}
\BIBentryALTinterwordspacing
A.~Krizhevsky, V.~Nair, and G.~Hinton. {Cifar-10 (canadian institute for
  advanced research)}. [Online]. Available:
  \url{http://www.cs.toronto.edu/˜kriz/cifar.html}
\BIBentrySTDinterwordspacing

\bibitem{Liu2017}
Y.~Liu, X.~Chen, C.~Liu, and D.~Song, ``{Delving into transferable adversarial
  examples and black-box attacks},'' \emph{International Conference on Learning
  Recognition(ICLR)}, 2017.

\bibitem{Madry2017}
\BIBentryALTinterwordspacing
A.~Madry, A.~Makelov, L.~Schmidt, D.~Tsipras, and A.~Vladu, ``{Towards deep
  learning models resistant to adversarial attacks},'' \emph{International
  Conference on Learning Recognition(ICLR)}, 2018. [Online]. Available:
  \url{https://arxiv.org/abs/1706.06083}
\BIBentrySTDinterwordspacing

\bibitem{Marcel2010}
\BIBentryALTinterwordspacing
S.~Marcel and Y.~Rodriguez, ``{Torchvision the machine-vision package of
  torch},'' \emph{Proceedings of the 18th ACM International Conference on
  Multimedia}, p. 1485–1488, 2010. [Online]. Available:
  \url{https://doi.org/10.1145/1873951.1874254}
\BIBentrySTDinterwordspacing

\bibitem{Papernot2017a}
N.~Papernot, P.~McDaniel, I.~Goodfellow, S.~Jha, Z.~Celik, and A.~Swami,
  ``{Practical black-box attacks against machine learning},'' \emph{ACM on Asia
  Conference on Computer and Communications Security (ASIA CCS)}, pp. 506--519,
  2017a.

\bibitem{Papernot2016b}
N.~Papernot, P.~McDaniel, S.~Jha, M.~Fredrikson, Z.~B. Celik, and A.~Swami,
  ``{The limitations of deep learning in adversarial settings},''
  \emph{Security and Privacy, 2016 IEEE European Symposium}, pp. 372--387,
  2016.

\bibitem{Papernot2016a}
N.~Papernot, P.~McDaniel, X.~Wu, S.~Jha, and A.~Swami, ``{Distillation as a
  Defense to Adversarial Perturbations Against Deep Neural Networks},''
  \emph{IEEE Symposium on Security and Privacy (SSP)}, 2016.

\bibitem{Selvaraju2017}
R.~Selvaraju, A.~Das, R.~Vedantam, M.~Cogswell, D.~Parikh, and D.~Batra,
  ``{Grad-CAM: Visual Explanations from Deep Networks via Gradient-based
  Localization},'' \emph{The IEEE International Conference on Computer Vision
  (ICCV)}, 2017.

\bibitem{Suya2020}
F.~Suya, J.~Chi, D.~Evans, and Y.~Tian, ``{Hybrid Batch Attacks: Finding
  Black-box Adversarial Examples with Limited Queries},'' \emph{USENIX Security
  Symposium}, 2020.

\bibitem{Szegedy2013}
\BIBentryALTinterwordspacing
C.~Szegedy, W.~Zaremba, I.~Sutskever, J.~Bruna, D.~Erhan, I.~Goodfellow, and
  R.~Fergus, ``{Intriguing properties of neural networks},''
  \emph{International Conference on Learning Recognition(ICLR)}, 2014.
  [Online]. Available: \url{https://arxiv.org/abs/1312.6199}
\BIBentrySTDinterwordspacing

\bibitem{Tramer2018}
F.~Tramer, A.~Kurakin, N.~Papernot, I.~Goodfellow, D.~Boneh, and P.~McDaniel,
  ``{Ensemble adversarial training: Attacks and defenses},''
  \emph{International Conference on Learning Recognition(ICLR)}, 2018.

\bibitem{Xu2019}
K.~Xu, S.~Liu, P.~Zhao, P.~Chen, H.~Zhang, Q.~Fan, D.~Erdogmus, Y.~Wang, and
  X.~Lin, ``{Structured Adversarial Attack: Towards General Implementation and
  Better Interpretability},'' \emph{International Conference on Learning
  Recognition(ICLR)}, 2019.

\end{thebibliography}

\begin{appendices}

\section{Computation Time of Experiments} 
\label{apd-experiment time summary}

\begin{table}[h!]
\caption{Summary of computation time for each experiment}
\centering
\begin{tabular}{l|c}
\toprule
Experiments & Duration \\ 
\midrule
Robustness of \rambo (Sec. \ref{sec:Robustness of Hybrid Methods})& 627 hrs\\ 
Benchmark on \hardupper $\&$ \easyupper sets (Sec. \ref{sec:benchmark on hard and non hard set}) & 275 hrs \\ 
Impact of the Starting Images (Sec. \ref{sec:sensitivity-to-starting-image}) & 38 hrs\\ 
Visual Explanation (Sec. \ref{sec:visual explanations}) & 5 hrs\\
Attack against a Defended Models (Sec. \ref{sec:Attack against a Defended Models}) & 281 hrs\\ 
Hyper-Parameters and Impacts (App. \ref{apd-hyper-parameters}) & 60 hrs\\
Validation on Balance Datasets (App. \ref{apd-Validation on balance datasets}) & 414 hrs\\
Untargeted Attack Validation (App. \ref{apd-untargeted attack}) & 126 hrs\\
\midrule
Total & 1826 hrs\\
\bottomrule
\end{tabular}

\label{table:computation time}
\end{table}

\section{Hyper-parameters and Impacts} \label{apd-hyper-parameters}
\noindent\textbf{Gradient Estimation:} The main hyper-parameter $n_\text{t}$ used in gradient estimation method is to control when the first component terminates and switches to \RSBCD. In practice, we keep track of query numbers executed and distortion between the source image and a crafted sample per iteration. This information is then used to determine distortion reduction rate $\Delta$ over $T$ queries. On \texttt{CIFAR10}, if applying HopSkipJump or Sign-OPT to the first component, $T = 500 ~or ~400$, respectively while on \texttt{ImageNet}, $T = 2000 ~or ~1000$, respectively.

\vspace{1mm}
\noindent\textbf{\RSBCD}: The hyper-parameters used are $n = 1$, 
initial $\delta =P_\text{i}(|\boldsymbol{x} - \boldsymbol{x}_\text{s}|)$, $m=1, \lambda=1.2, \epsilon_r=0.01, \epsilon_s=0.01$ for GradEstimation, $\epsilon_s=0.01$ for \RSBCD, $T = 500$ and $P_\text{i}=P_\text{100}$. For the \textit{larger} dataset, \texttt{ImageNet}, the changes are: $m=16, \lambda=2, \epsilon_r=0.1, \epsilon_s=1$ for GradEstimation, $\epsilon_s=0.1$ for \RSBCD, $T=1000$ and $P_\text{i}=P_\text{50}$.

\noindent\textbf{The impact of parameter $\lambda$:~} The key parameter that may influence \RSBCD is $\lambda$ because it controls the step size (or perturbation magnitude $\delta$) for each cycle (see line 28 in Algorithm \ref{algo2}). For example, $\lambda$ is used to determine the step from $x^{(4)}$ to $x^{(5)}$ in Fig. \ref{fig:toy example - hybrid}. If $\lambda$ is small, $\delta$ reduces slightly and thus remains relatively large after each cycle. Consequently \RSBCD takes large movements that are likely to yield large magnitude adversarial examples and/or miss the optimal solution. Alternatively, it may cross the decision boundary into an undesired class (source image class in a targeted attack).

\begin{figure}[htp]
    \begin{center}
        \includegraphics[width=0.6\linewidth,trim={0 0 0 0},clip]{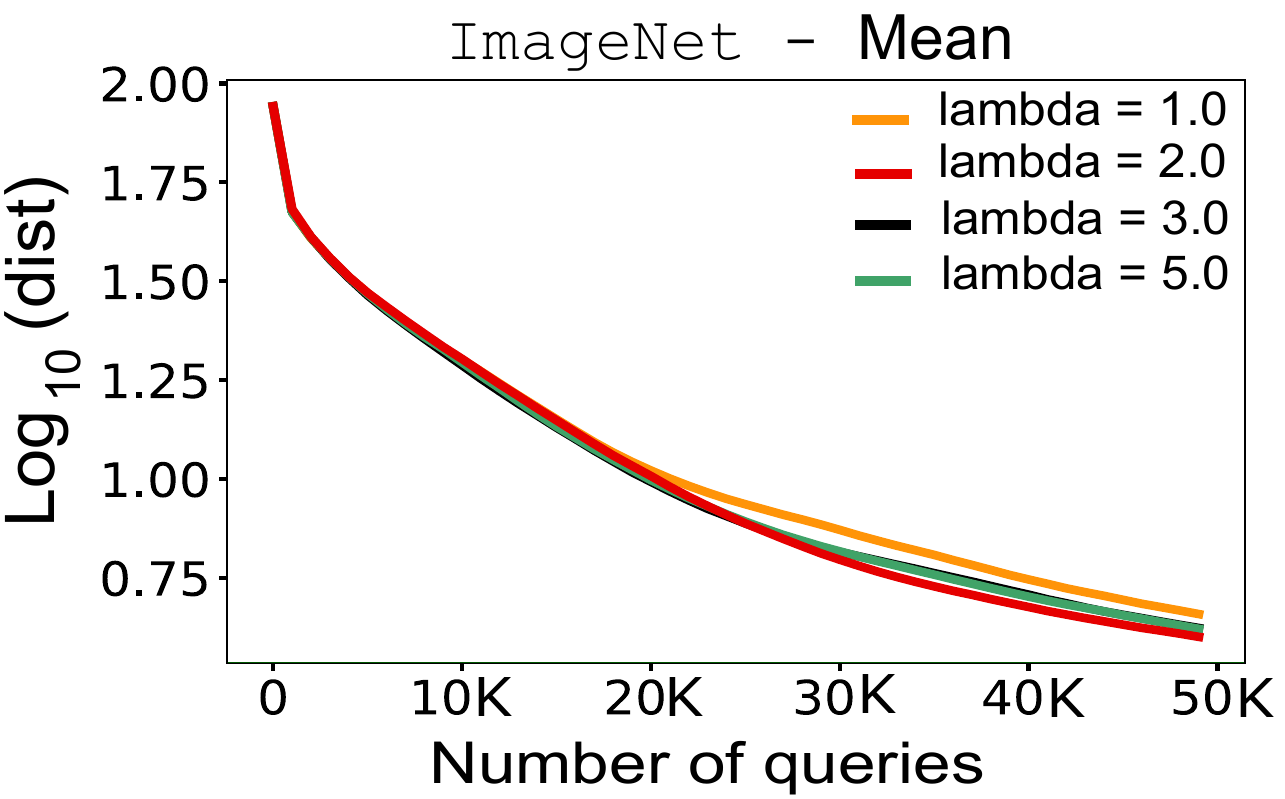}
        \caption{A comparison between \rambo with different values of $\lambda$ on 100 source and target class sample pairs selected from \texttt{ImageNet}.}
        \label{fig:hyper-parameter-lambda}
    \end{center}
    \vspace{-3mm}
\end{figure}

In contrast, if $\lambda$ is large, \RSBCD takes finer steps to yield adversarial samples whilst moving towards the source image and likely stay in the desired class (target class in a targeted attack). Nevertheless, the empirical result with 100 pairs of source and target class images on \texttt{ImageNet} shown in Fig. \ref{fig:hyper-parameter-lambda} 
illustrates that the overall performance of \rambo is not greatly affected by $\lambda$ and at $\lambda=2$, \rambo achieves the best performance. 

\section{Proposed Robustness Evaluation Protocol} \label{apd-Robustness Evaluation Protocol}

\begin{figure}[htp]
    \begin{center}
        \includegraphics[width=0.6\linewidth,trim={0 0 0 1cm},clip]{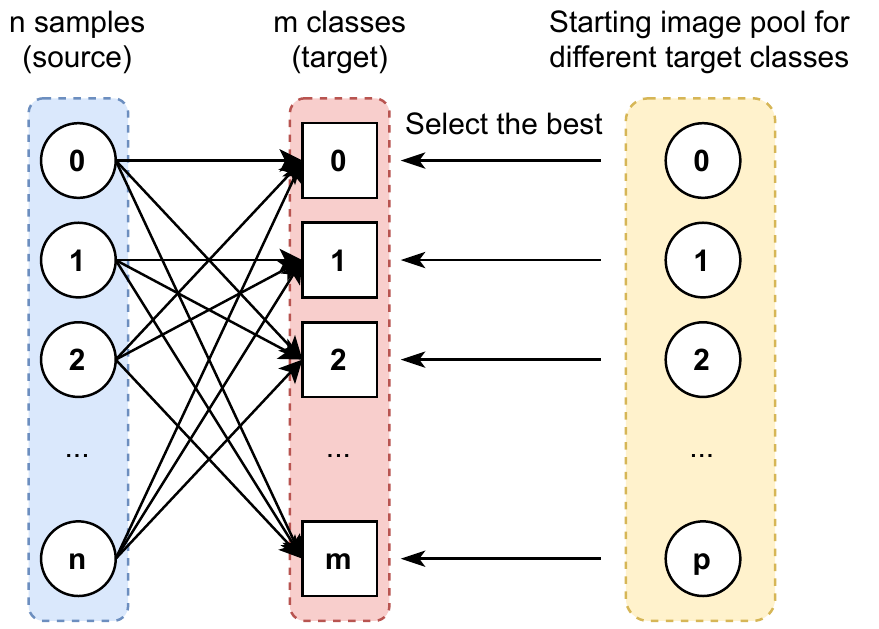}
        \caption{The proposed evaluation protocol for assessing robustness under an exhaustive evaluation setting. In this mode, each sample from a dataset with size of $n$ is evaluated to obtain an adversarial example for that sample capable of flipping its predicted label to $m$ different target classes from that dataset. For each attack,a starting image is selected from a pool of $p$ starting images.}
        \label{fig:robustness evaluation protocol}
    \end{center}
    \vspace{-3mm}
\end{figure}

An attack method is mounted to change the true prediction of the DNN from its ground truth label for a given source sample image to each of the different different target classes. For \texttt{CIFAR10} with ten classes, an attack method selects each of the 1000 test set samples for a given class as a source image and attempts to find an adversarial example for each of the other target classes (of which there are 9). Consequently, we evaluate 90,000 pairs of source and starting images. Since there is no effective method to choose a starting image from a target class, for a fair evaluation, we apply the same protocol used in~\cite{Cheng2019b, Cheng2020} to initialize an attack for each method. We execute each attack with a query budget of 50,000 queries. Then we identify \hard cases of each attack method against the victim model (detailed in Section~\ref{experiment setting}). This protocol can be generalized to other datasets by choosing $n$ samples and $m$ different target classes from that dataset where each target class has its own starting image as shown in Fig.~\ref{fig:robustness evaluation protocol}.
\begin{figure}[htp]
    \begin{center}
        \includegraphics[width=0.5\linewidth,trim={0 0 0 0},clip]{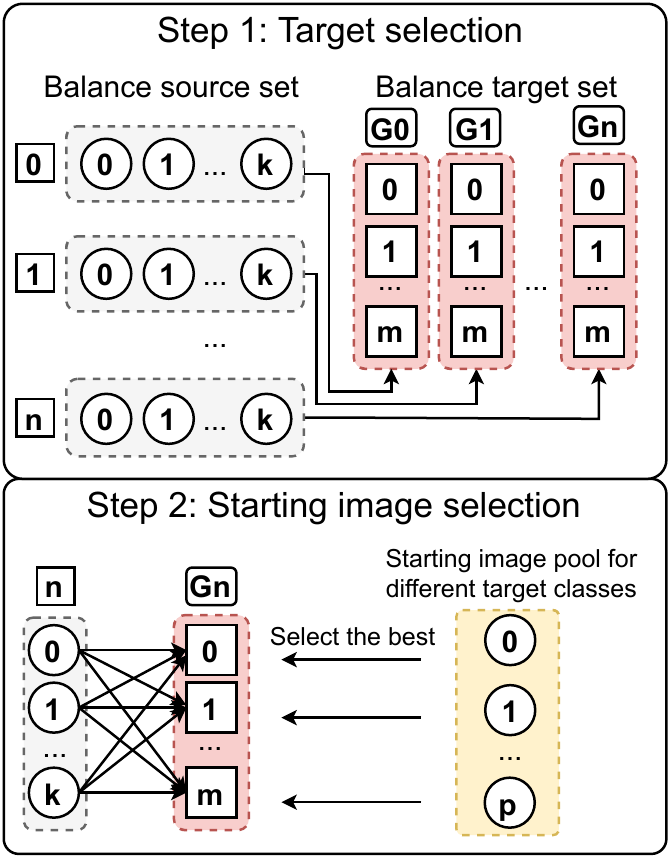}
        \caption{The proposed evaluation protocol requires a balance dataset including $n$ source classes and a balance target set comprising of $n$ corresponding groups. On balance source set, all source classes have an equal number of samples ($k$) while all $n$ corresponding groups have an equal number of target class ($m$). These target classes are different within a group but can be repeated in other groups. From these groups $G_n$, a starting image is selected from a pool of $p$ starting images.}
        \label{fig:evaluation protocol}
    \end{center}
    \vspace{-3mm}
\end{figure}

\begin{table}[t]
\centering
\caption{Summary comparison among attacks with \rambo on small and large scale balance datasets.}

\label{table:balance set}
\resizebox{1.0\linewidth}{!}{
\begin{tabular}{c|c|cccc|cccc} 
\toprule
Query                & \multirow{2}{*}{Methods} & \multicolumn{4}{c}{CIFAR10}                                                        & \multicolumn{4}{c}{ImageNet}                                                     \\ 
\cline{3-10}
budget               &                          & Mean           & Std            & Median         & ASR($\epsilon$=0.3)             & Mean           & Std           & Median        & ASR                             \\ 
\midrule
\multirow{5}{*}{25K} & Boundary                 & 0.674          & 0.654          & 0.499          & 22.6$\boldsymbol{\%}$           & 31.80          & 18.43         & 32.88         & 5.5$\boldsymbol{\%}$            \\
                     & HopSkipJump              & 0.507          & 0.748          & 0.296          & 50.8$\boldsymbol{\%}$           & 11.91          & 8.39          & 10.87         & 51.4$\boldsymbol{\%}$           \\
                     & Sign-OPT                 & 0.526          & 0.754          & 0.286          & 53.6$\boldsymbol{\%}$           & 14.21          & 11.52         & 9.81          & 46.3$\boldsymbol{\%}$           \\ 
\cmidrule{2-10}
                     & \textbf{RamBo.} (HSJA)                   & \textbf{0.336} & \textbf{0.218} & 0.283          & 54.0$\boldsymbol{\%}$           & 11.33          & \textbf{8.0}  & \textbf{8.62} & 53.1$\boldsymbol{\%}$           \\
                     & \textbf{RamBo.} (SOPT)                   & 0.363          & 0.359          & \textbf{0.282} & \textbf{54.1}$\boldsymbol{\%}$  & \textbf{11.25} & 9.47          & 9.62          & \textbf{57.5}$\boldsymbol{\%}$  \\ 
\midrule
\multirow{5}{*}{50K} & Boundary                 & 0.399          & 0.404          & 0.319          & 45.2$\boldsymbol{\%}$           & 23.73          & 15.65         & 20.71         & 16.6$\boldsymbol{\%}$           \\
                     & HopSkipJump              & 0.460          & 0.683          & 0.273          & 55.3$\boldsymbol{\%}$           & 7.09           & 5.11          & 4.87          & 82.0$\boldsymbol{\%}$           \\
                     & Sign-OPT                 & 0.420          & 0.562          & 0.267          & 59.1$\boldsymbol{\%}$           & 7.79           & 7.84          & 5.87          & 73.3$\boldsymbol{\%}$           \\ 
\cmidrule{2-10}
                     & \textbf{RamBo.} (HSJA)                   & \textbf{0.300} & \textbf{0.178} & \textbf{0.260} & 59.9$\boldsymbol{\%}$           & \textbf{4.80}  & \textbf{3.70} & 3.92          & \textbf{93.1}$\boldsymbol{\%}$  \\
                     & \textbf{Rambo.} (SOPT)                   & 0.306          & 0.193          & 0.261          & \textbf{60.11}$\boldsymbol{\%}$ & 5.02           & 4.57          & \textbf{3.84} & 92.3$\boldsymbol{\%}$           \\
\bottomrule
\end{tabular}
}
\end{table}

\section{Proposed Validation Protocol for Balanced sets and Results on Non-hard Sets} \label{apd-Validation on balance datasets}

\begin{figure}[htp]
    \begin{center}
        \includegraphics[scale=0.35]{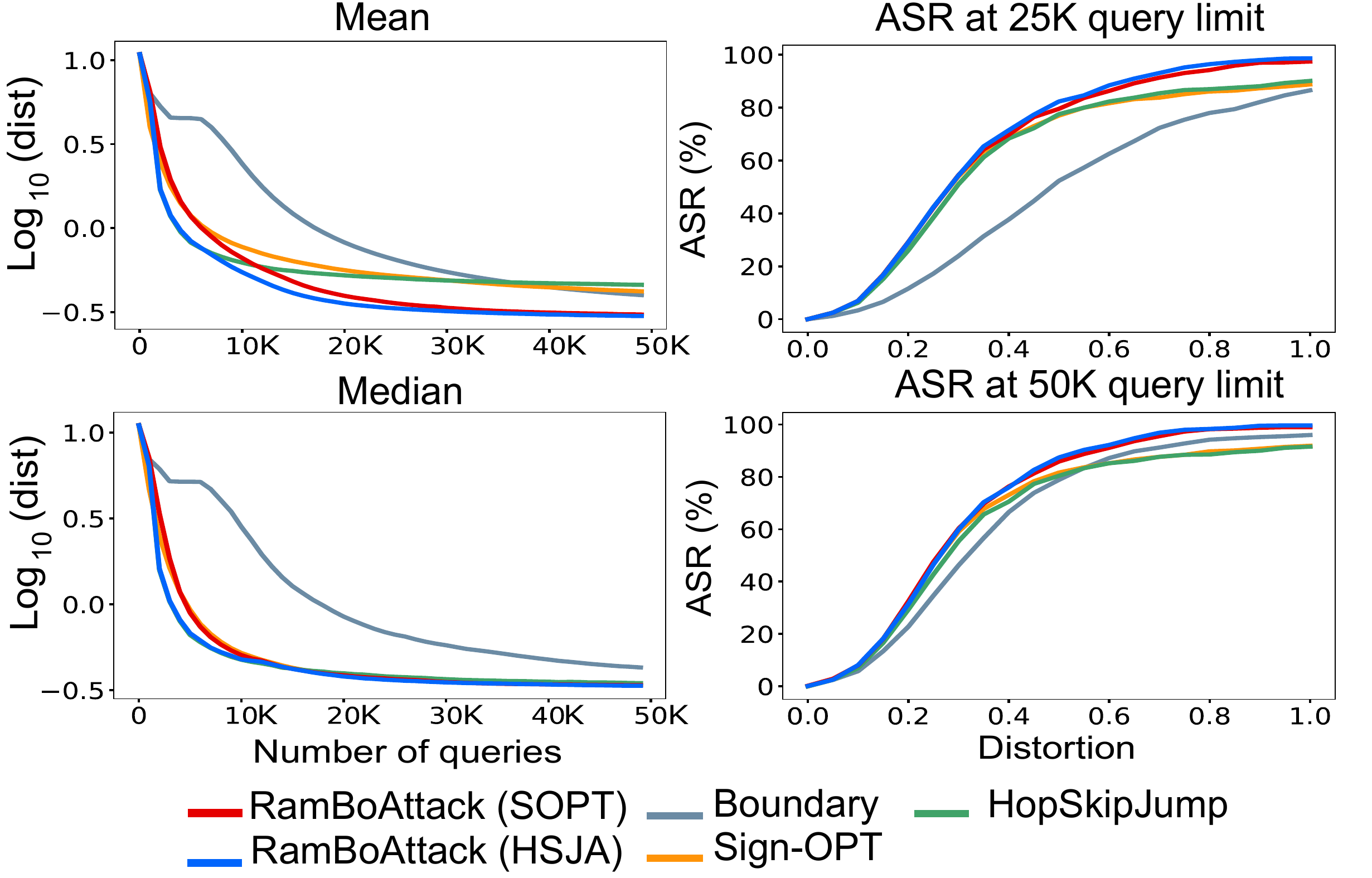}
        \caption{A comparison between three current state-of-the-art attacks and \rambo on a balance set selected from \texttt{CIFAR10}.}
        \label{fig:small scale balance set}
    \end{center}
    \vspace{-3mm}
\end{figure}

\begin{figure}[htp]
    \begin{center}
        \includegraphics[scale=0.45]{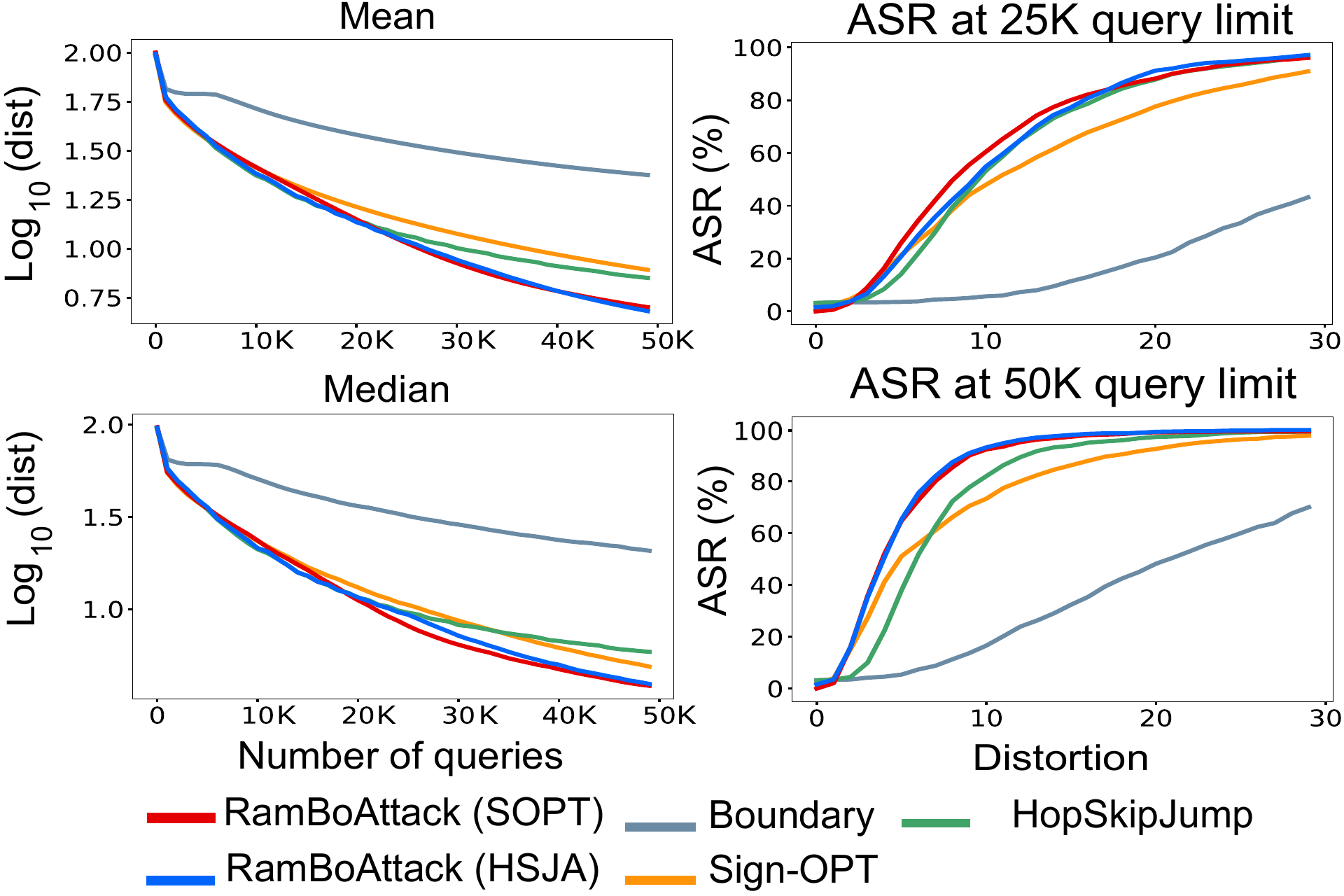}
        \caption{A comparison between three current state-of-the-art attacks and \rambo on a large scale balance set selected from \texttt{ImageNet}.}
        \label{fig:large scale balance set}
    \end{center}
    \vspace{-3mm}
\end{figure}
\noindent\textbf{Evaluation protocol.} The second research question highlights a need to evaluate the overall performance of various blackbox attacks under decision-based settings reliably. On \texttt{CIFAR10}, most previous works propose to choose a random evaluation set with randomly sampled images with label $y$ and select a random target label $\Tilde{y}$ \cite{Cheng2020} or set $\Tilde{y}=(y+1)$ mod 10~\cite{Brendel2018, Brunner2019, Cheng2019b}. Nonetheless, these selection schemes may lead to an imbalanced dataset that is insufficient to evaluate the effectiveness of the attack since it may lack the so called \hard cases that occur more frequently with specific pairs of classes. As a result, it may lead to a bias in evaluation results and fail to highlight potential weaknesses of an attack. Consequently, were were motivated to propose a more robust and reliable evaluation protocol and illustrate it in Fig. \ref{fig:evaluation protocol}.

\noindent\textbf{On balance sets:~}A balance set comprises of a balanced source set and a balanced target set. Both sets are composed of $N$ different source classes and $N$ corresponding groups. Each group is composed of $m$ different target classes and all source and target classes are randomly chosen from all classes of a test set. In addition, all target classes are different within a group but can be repeated in other groups. Each source class has $n$ samples selected randomly from a test set. Adversaries may have one or several images from each target class and select one to initialize an attack. Each attack method aims to craft an adversarial example for every selected sample from each source class and flip its true prediction towards every target class given in the corresponding group of balanced target set. The total number of evaluation pairs is $N \times n \times m$. For instance, every sample of source class $i$ ($\text{img: }i_1, i_2, \cdots, i_n$) is  flipped towards each target class ($\text{class: }i_1, i_2, \cdots, i_m$) in the corresponding group $i$ (see Fig.~\ref{fig:evaluation protocol}). 

\vspace{1mm}
\noindent\textbf{Balanced Set with \texttt{CIFAR10}.} It is simple to carry out a comprehensive evaluation over all classes, so we choose N=10, n=10 and m=9. In addition, to demonstrate the query efficiency and effectiveness of each attack, we employ a query budget of 25,000 and 50,000 across all experiments. \rambo obtain slightly better median and mean distortion than HopSkipJump and Sign-OPT at 25K and 50K, as shown in Table \ref{table:balance set}. On the standard deviation metric used to measure  distortion variance across an evaluation set, our \rambo outperform Boundary, HopSkipJump and Sign-OPT at query limit of 25K and 50K. In order words, our attack performs robustly  across the evaluation set. 

\vspace{1mm}
\noindent\textbf{Balanced Set with \texttt{ImageNet}.} \texttt{ImageNet} has 1000 distinct classes, hence carrying out a comprehensive evaluation like on \texttt{CIFAR10} requires huge computing resources and time. Therefore, we choose N=200, n=1 m=5 and limit the query budget to 25,000 and 50,000. The average distortion (on a $\log_{10}$ scale) against the queries and attack success rate (ASR) at 25K and 50K query budgets achieved by \rambo is better than Boundary, Sign-OPT and HopSkipJump attacks as seen in shown in Fig. \ref{fig:large scale balance set}. As shown in Table \ref{table:balance set}, on average distortion metric, \rambos obtain better result and achieve significantly smaller standard deviation of distortion overall .

\begin{figure}[htp]
    \begin{center}
        \includegraphics[scale=0.3]{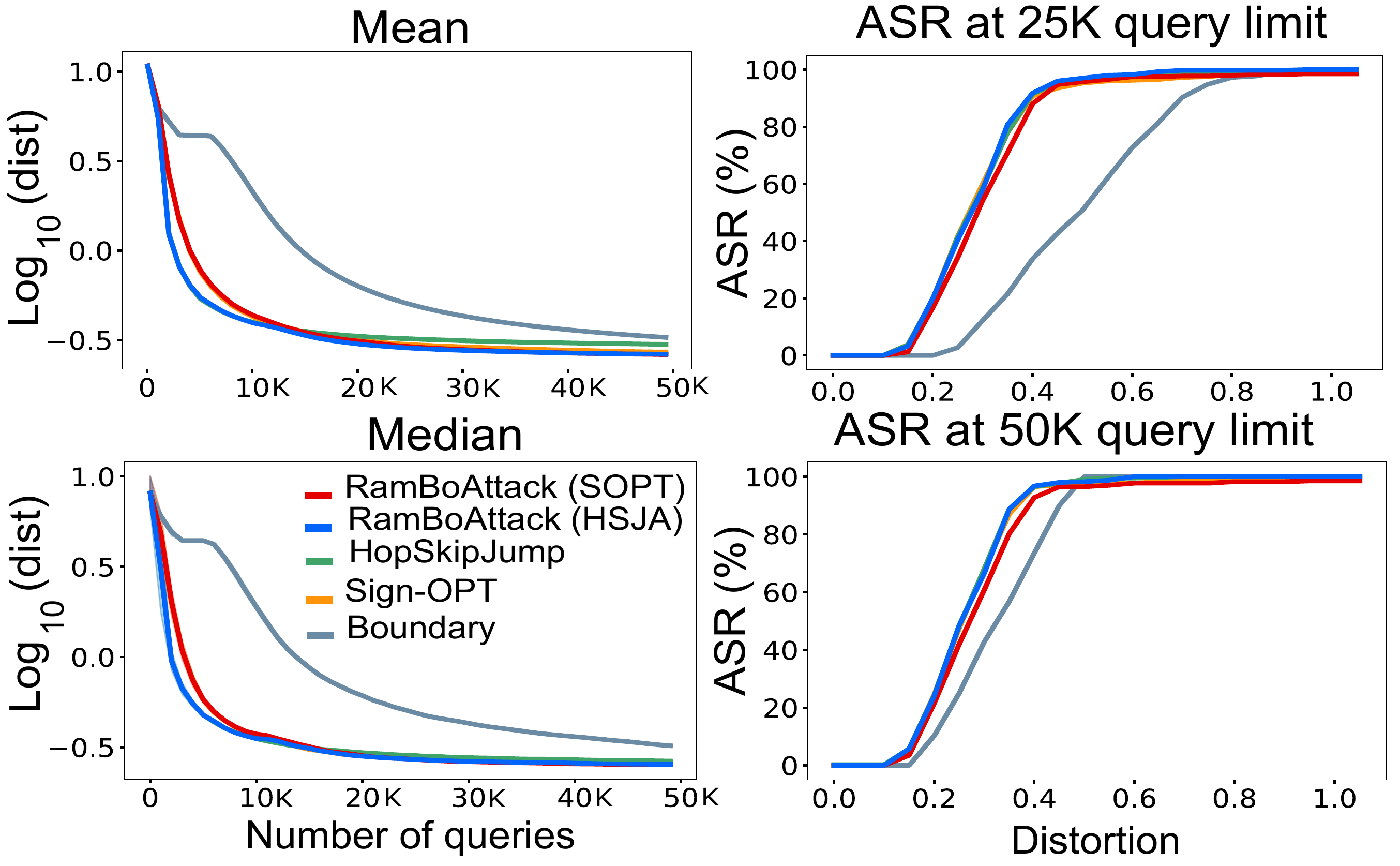}
        \caption{A comparison between three current state-of-the-art attacks and \rambo on a \easy set C selected from \texttt{CIFAR10}. In \easy cases, we perform comparably.}
        \label{fig:cifar10 non hard set}
    \end{center}
    \vspace{-3mm}
\end{figure}

\begin{figure}[htp]
    \begin{center}
        \includegraphics[scale=0.3]{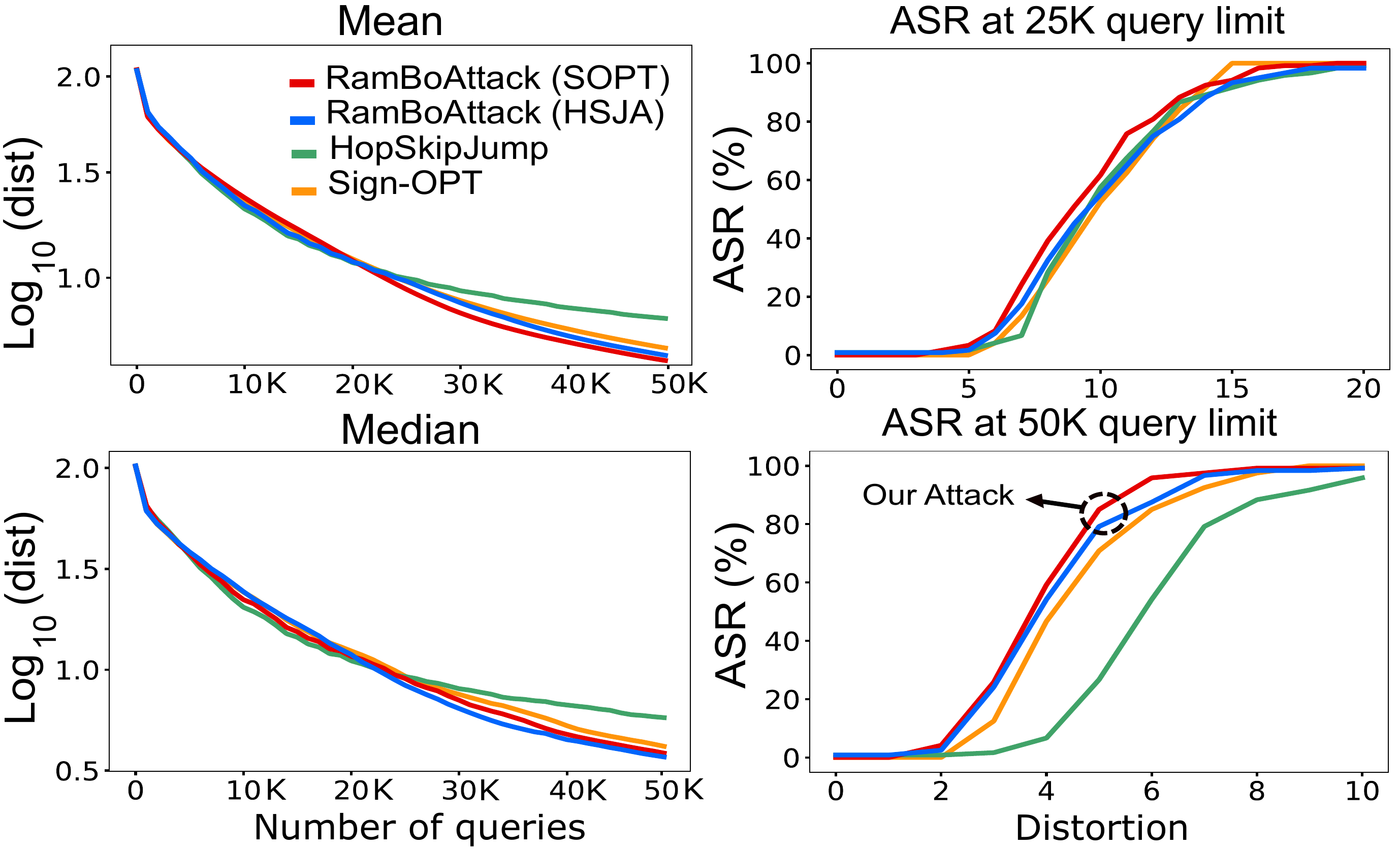}
        \caption{A comparison between three current state-of-the-art attacks and \rambo on a \easy set selected from \texttt{ImageNet}. In \easy cases, \rambos improve attack performance by yielding more effective adversarial examples notable in ASR results.}
        \label{fig:ImageNet non hard set}
    \end{center}
    \vspace{-3mm}
\end{figure}

\noindent\textbf{On \easy sets:}
In this section, we evaluate the performance of SignOPT, HopSkipJump and our \rambos on both \texttt{CIFAR10} and \texttt{ImageNet} \easy set. The common \easyset C drawn from \texttt{CIFAR10} for all methods is composed of 400 \easy sample pairs. They are selected such that a distortion between a source image and its adversarial example found after 50,000 is smaller or equal 0.6. Likewise, a \easyset from \texttt{ImageNet} is composed of $120$ \easy sample pairs and the distortion threshold to select these is is 7. Fig.~\ref{fig:cifar10 non hard set} and~\ref{fig:ImageNet non hard set} show that our attack has comparable performance to SignOPT and HopSkipJump on \texttt{CIFAR10} \easy subsets whilst demonstrating improved attack performance by yielding more effective adversarial examples, especially with a 50K query budget, as seen in the higher attack success rates obtained by \rambos.

\section{Untageted Attack Validation} \label{apd-untargeted attack}

\begin{figure}[htp]
    \begin{center}
        \includegraphics[scale=0.45]{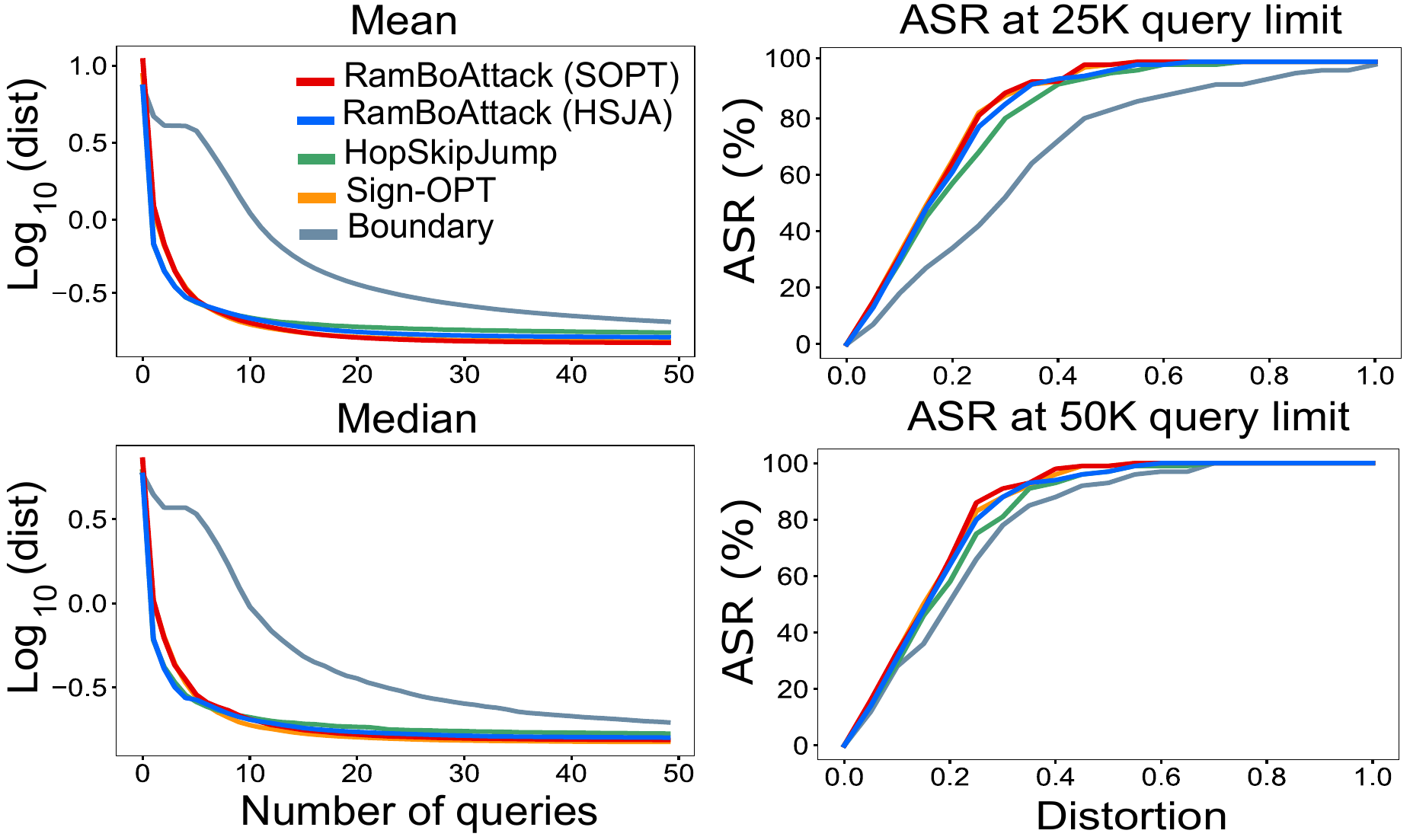}
        \caption{Comparing between three current state-of-the-art attacks and \rambo on the balance set selected from \texttt{CIFAR10} under untargeted setting.}
        \label{fig:untargeted CIFAR10}
    \end{center}
    \vspace{-3mm}
\end{figure}

\begin{figure}[htp]
    \begin{center}
        \includegraphics[scale=0.45]{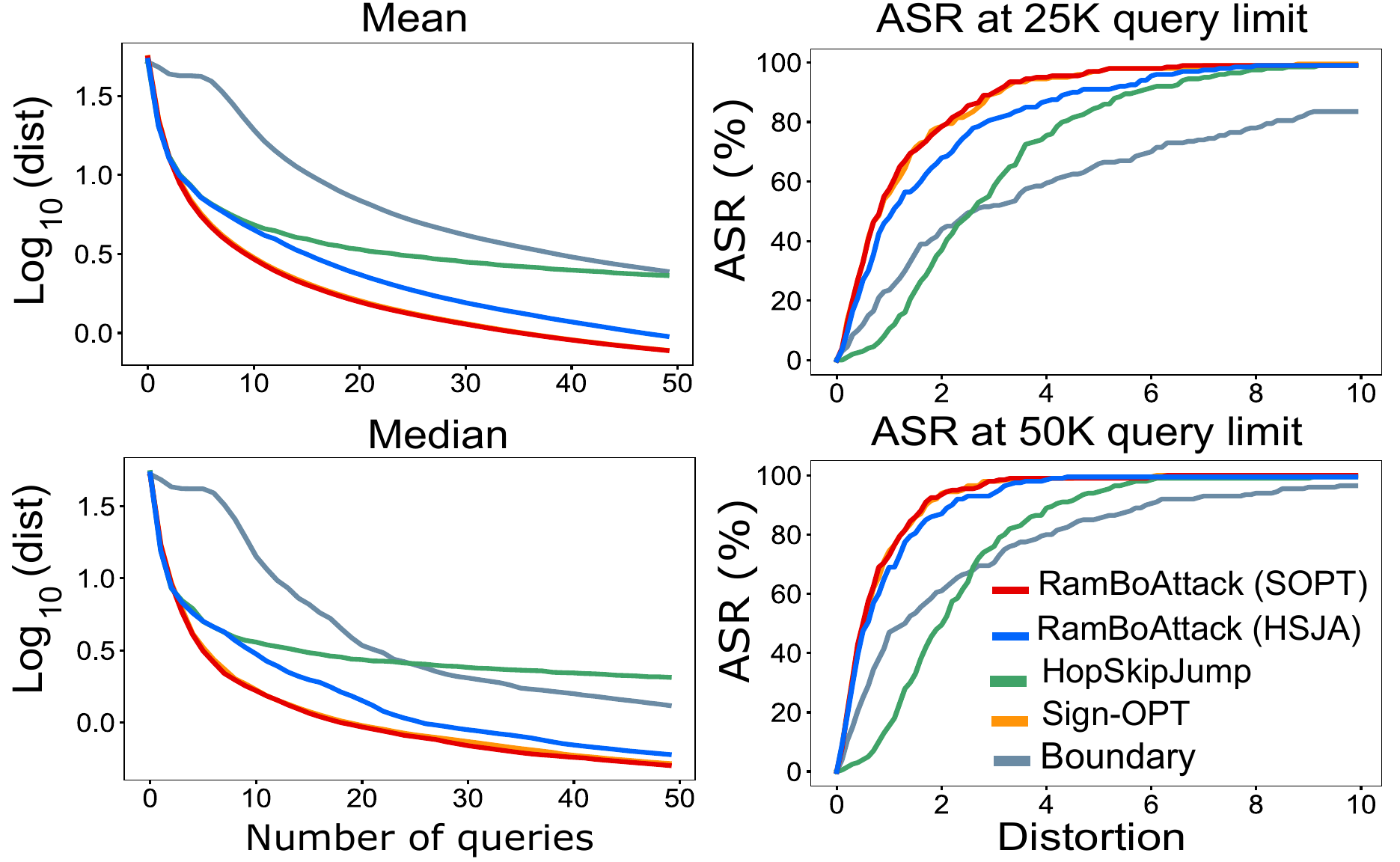}
        \caption{Comparing between three current state-of-the-art attacks and \rambo on the balance set from \texttt{ImageNet} under untargeted setting.}
        \label{fig:untargeted ImgNet}
    \end{center}
    \vspace{-3mm}
\end{figure}

Here, we evaluate our \rambo and other state-of-the-art attacks on two different balanced sets from \texttt{CIFAR10} and \texttt{ImageNet} as described in Appendix~\ref{apd-Validation on balance datasets} under an untargeted scenario. for completeness. First, on the balance set from \texttt{CIFAR10}
, our attacks can achieve comparable performance with Sign-OPT and HopSkipJump and obtain approximately 97$\%$ success rate at a distortion of 0.5 on a 25K query budget (see Fig.~\ref{fig:untargeted CIFAR10}); however, our attack method outperforms Boundary attack. In contrast, on the balance set selected from \texttt{ImageNet}, we observe that our methods can achieve comparable performance with Sign-OPT but outperform HopSkipJump and Bourndary attack as shown in Fig.~\ref{fig:untargeted ImgNet}.     

\section{Impact of Starting Images} \label{apd:Impact of Starting Images}
\begin{figure}[ht]
    \begin{center}
        \includegraphics[scale=0.54]{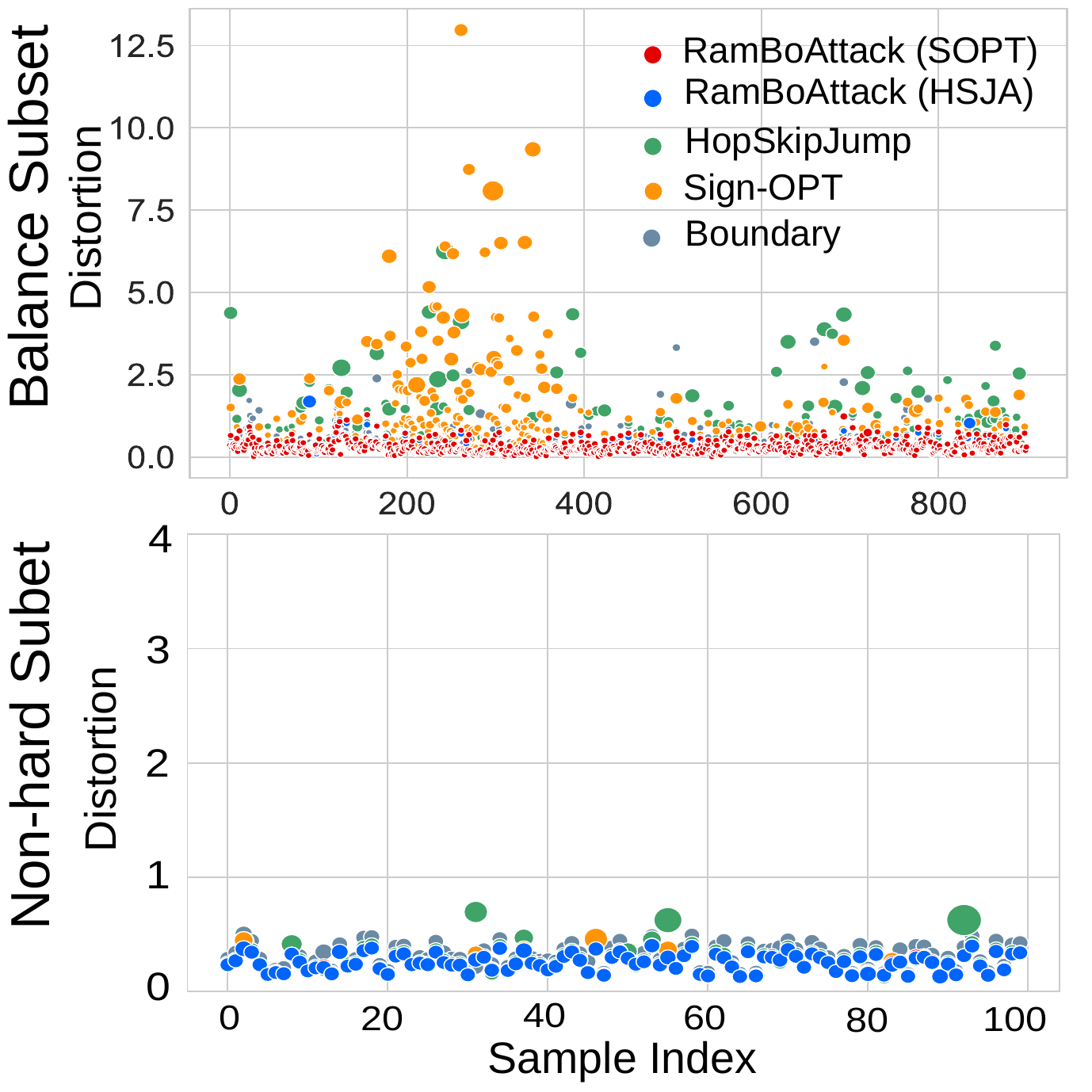}
        \caption{An illustration of the sensitivity of different attacks to various chosen starting images. Size of each circle denotes standard deviation and y-axis indicates the mean distortion. The results are from the \texttt{CIFAR10} balance set and a \easy subset from \easy set C. Compared with Boundary, Sign-OPT and HopSkipJump attacks, our \rambos are \textbf{\textit{much less sensitive to the choice of starting image}} in general. On \easy cases, all of attacks can achieve comparable results. Hence our attack is demonstrably more robust.}
        \label{fig:one vs various starting image-balance set}
    \end{center}
    \vspace{-3mm}
\end{figure}
In this section, we first compose a \easy subset with 100 random \easy sample pairs selected from \easy set C. We also compose a balance subset from the balance set described in Appendix \ref{apd-Validation on balance datasets}. We then evaluate our \rambo, Sign-OPT, HopSkipJump and Boundary attack on these subsets. To conduct this experiment, for every source image and each of its target classes, we randomly select 10 different starting images and these attacks are executed with a query budget of 50K. We calculate the mean and standard deviation of distortion for each sample to measure the robustness of each attack to yield adversarial examples for each source image and target class pair. 

In Fig. \ref{fig:one vs various starting image-balance set}, size of each bubble denotes the standard deviation while the y-axis indicates mean distortion value. We can see that, on the \easy subset, the \rambos are able to achieve comparable result to all of the state-of-the-art methods. On the balance subset, our \rambos can achieve significantly less variance (smaller bubbles) at lower distortions while most results achieved by Sign-OPT, HopSkipJump and Boundary indicate larger variance (larger bubbles) and higher distortions. Consequently, our \rambos are more robust than Sign-OPT and HopSkipJump and less sensitive to the chosen starting image.

\section{Attack against Defended Models}\label{apd-defended model}

\begin{figure}[ht]
    \begin{center}
        \includegraphics[scale=0.4]{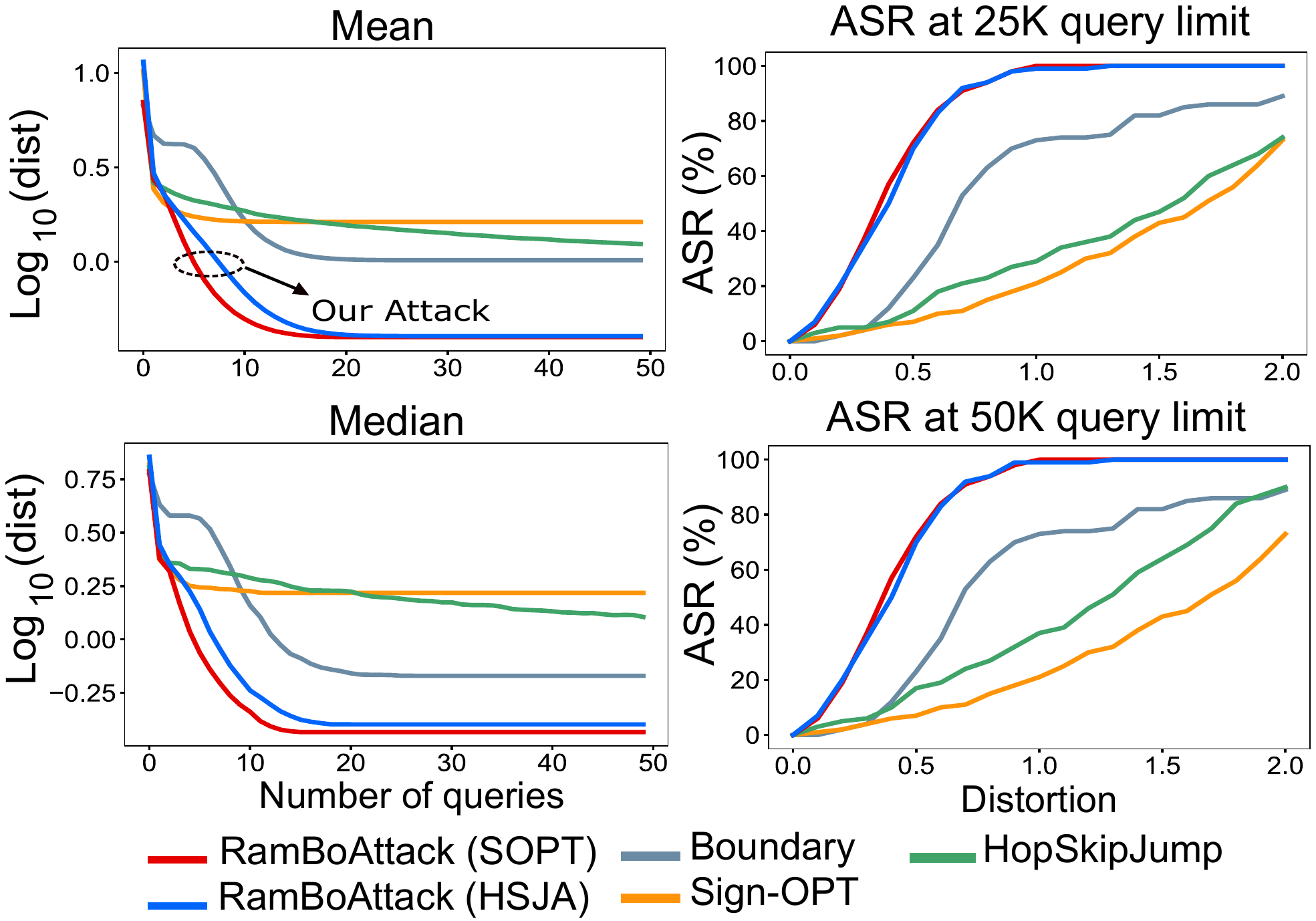}
        \caption{Performance comparison between different state-of-the-art attacks and \rambo against a region-based classifier on \texttt{CIFAR10}. \rambo outperforms other blackbox attacks and is able to craft \textit{significantly more effective} adversarial examples of lower distortion against the defense method as seen by the higher ASR results against the defended models from \rambo across all of the evaluations.}
        \label{fig:regional base classifier}
    \end{center}
    \vspace{-5mm}
\end{figure}

\begin{figure}[htp]
    \begin{center}
        \includegraphics[width=1.0\linewidth]{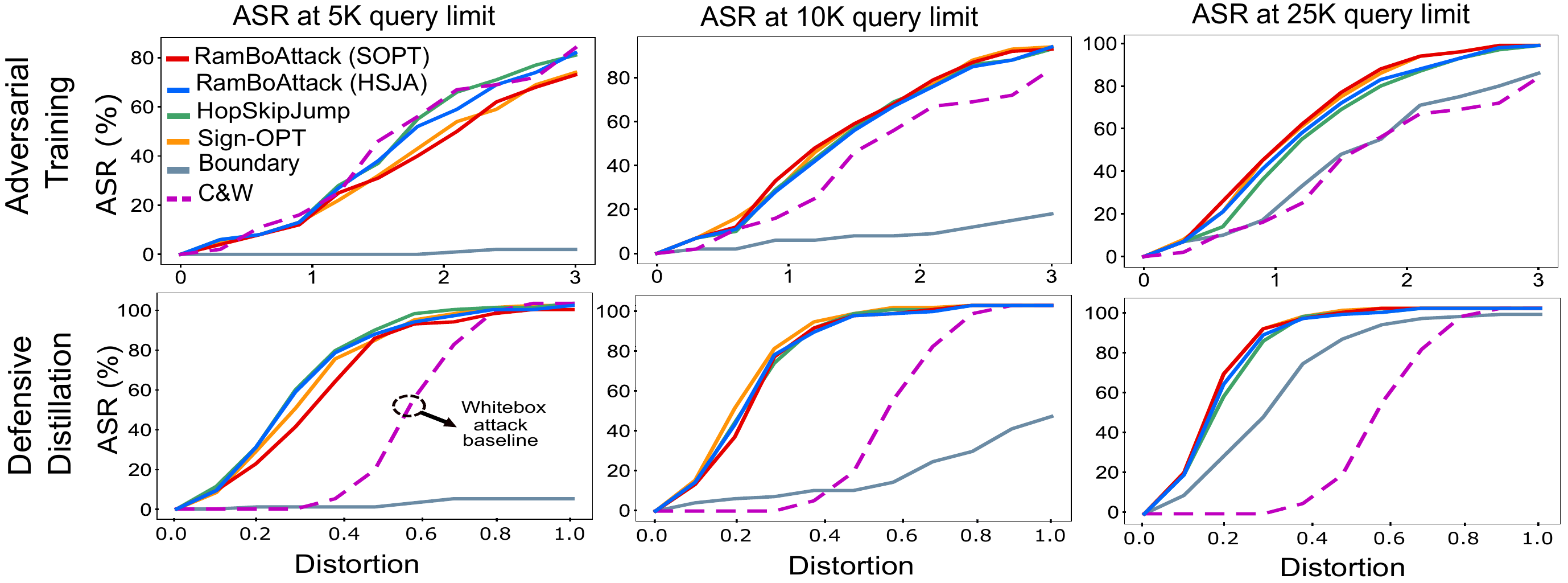}
        \caption{ASR comparison between white-box (\textit{employed as a baseline}) and current decision-based attacks versus our \rambo against Adversarial Training model and Defensive Distillation on \texttt{CIFAR10} (using the balanced set). Interestingly, \rambos are more effective than white-box attack method baseline, and are slightly more robust under different query settings when compared to other decision-based blackbox attacks.}
        \label{fig:ASR versus distortion ATModel}
    \end{center}
    \vspace{-3mm}
\end{figure}

In this section, we illustrate the results that we briefly mention in Section \ref{sec:Attack against a Defended Models}. Fig. \ref{fig:regional base classifier} shows that the average and median distortion (on a $\log_{10}$ scale) achieved by \rambos are significantly lower than BA, Sign-OPT and HopSkipJump. In addition, our attack  outperform others in terms of attack success rate (ASR) at 25K and 50K query budgets---i.e. achieves \textit{higher} ASR on defended models under different query budgets and distortion thresholds. Based on these results, \textit{we observe our attack to be robust than exiting attacks when mounting an attack against region-based classifiers}. 

The reason for this is that existing attack methods need to follow the decision boundary where region-based classifiers are capable of correcting its prediction by uniformly generating a large amount of data points at random and returning the most frequent predicted label. This capability of region-based classifiers prevents binary search in Sign-OPT and HopSkipJump from specifying the boundary exactly and results in noisy and coarse boundary estimations that cause all attack methods aiming to walk along the boundary fail to estimate a useful gradient direction. Nevertheless, our \rambos are able to break this defense mechanism because the core component, \RSBCD, is a derivative-free optimization that does not need to determine the boundary and estimate a gradient direction to descend. 

\subsection{Results}
Fig. \ref{fig:ASR versus distortion ATModel} shows the attack success rate (ASR) at different distortion levels and query limits for various attack methods against an adversarially trained model and defensive distillation model. Particularly, for adversarial training, our \rambos can achieve comparable performance with Sign-OPT and HopSkipJump while outperforming Boundary attack within the query limits of 5K, 10K or 25K. In addition, we compare the performance of our attack at different query budgets with the whitebox attack---C$\&$W---\textit{used as a baseline for comparison}. Notably, we do not execute C$\&$W attack at different query setting because it is a whitebox method and use the best result produced by this attack. 

We observe that our attacks are able to obtain a comparable performance with the C$\&$W attack at the 5K query budget. When the query limit is up to 10K and higher, our \rambos outperform the whitebox C$\&$W baseline attack method. Nevertheless, Adversarial Training is still effective at reducing the ASR achieved by our method, even with a 25K query budget. Success falls from around 99$\%$ (see Fig.~\ref{fig:untargeted CIFAR10}) to approximately 43$\%$ (see Fig.~\ref{fig:ASR versus distortion ATModel}) at a distortion of 1.0 ($l_2$ norm). Similarly, at a distortion of 0.3, the ASR decreases from about 60$\%$ (see Fig.~\ref{fig:untargeted CIFAR10}) to approximately 10$\%$ (see Fig.~\ref{fig:ASR versus distortion ATModel}). However, what we can observe is that as the distortion increases, the attack is more effective. This is expected because the attack budget of the adversary is increased above and beyond the budget used for generating the adversarial examples used for building the adversarially trained model.

Likewise, for defensive distillation, our \rambos can achieve comparable performance with Sign-OPT and HoSkipJump whilst  outperforming Boundary attack and C$\&$W whitebox baseline attack at different query budgets. These results confirm the results and findings presented in~\cite{Chen2020}.

\begin{figure}[htp]
    \begin{center}
        \includegraphics[scale=0.36]{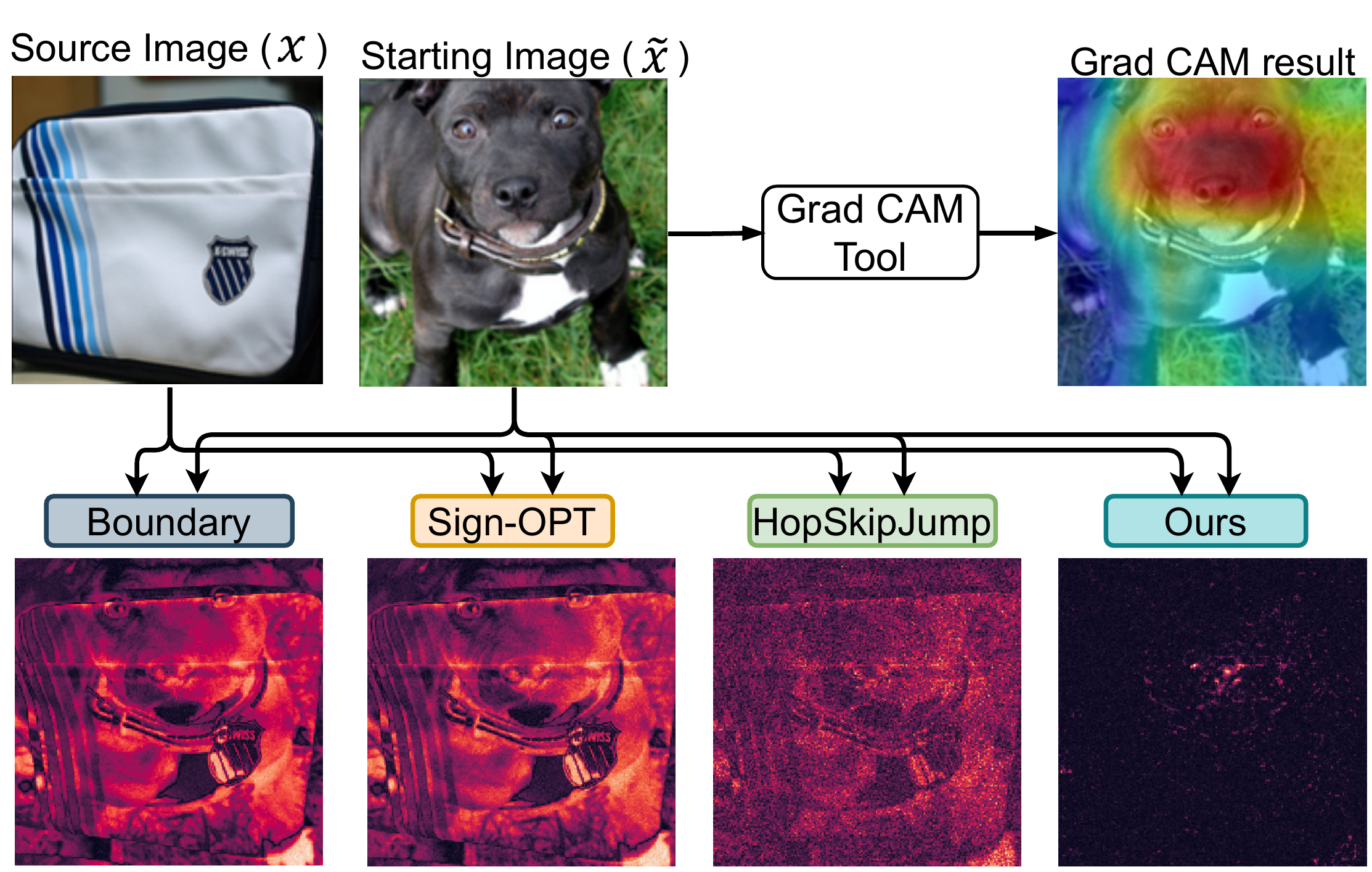}
        \caption{Grad-CAM tool visualizes salient area of the starting image \texttt{Staffordshire bull terrier}. Perturbation heat map (PHM) visualizes the normalized perturbation magnitude at each pixel. It shows that the perturbation yielded by \rambo is able to concentrate on salient areas illustrated by GRAD-CAM even though \rambo does not exploit the knowledge of salient regions to perturb.}
        \label{fig:apd-GradCAM-visual}
    \end{center}
    \vspace{-3mm}
\end{figure}

\begin{figure*}[ht]
    \begin{center}
        \includegraphics[width=0.9\linewidth,trim={0 0 0 0.1cm},clip]{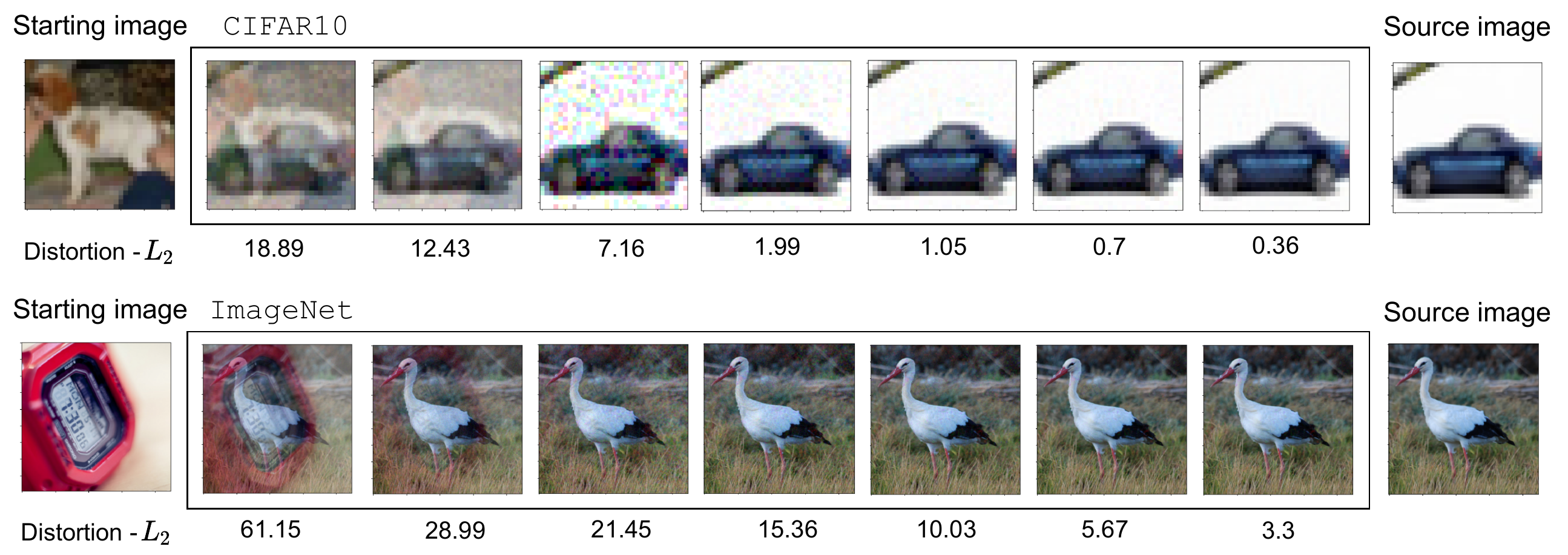}
        \caption{An illustration of different distortion levels produced by \rambo. The first row demonstrates an example from \texttt{CIFAR10} with a starting image of a \texttt{dog} gradually perturbed until it is similar to the source image \texttt{car}---the adversarial example. The bottom row demonstrates an example from \texttt{ImageNet} with is a starting image of a \texttt{digital watch} gradually perturbed until it is similar to the source image \texttt{white stork}---the adversarial example.}
        \label{fig:apd-Distortion level visualization}
    \end{center}
    \vspace{-3mm}
\end{figure*}

\subsection{C\&W Attack Configuration and Results Collection}\label{apd-CW Attack Configuration and Results Collection}
For clarity, here we describe the configuration used for the C$\&$W attack, the C$\&$W execution strategy, results collection for the C$\&$W attack and blackbox attacks.

For the C$\&$W attack, we adopt the PyTorch implementation of the C$\&$W method used in \cite{Cheng2019b,Cheng2020}. In their implementation, they use a learning rate of 0.1 and 1000 iterations for all evaluations (see published code). To search for an adversarial example for an image, the method performs a binary search step to find a relevant constant $c$ within a range from 0.01 to 1000 \textit{until a successful attack is achieved}. With this configuration, the C$\&$W attack is run once to always yield an adversarial example for every instance. We record the distortion of the adversarial example found.

\noindent\textbf{C$\&$W Results Collection.~}To construct ASR vs. distortion results, at different distortion thresholds: i)~we compute the number of source images in the evaluation set meeting a given distortion threshold (along the x-axis); ii)~then divide this by the total number of images in the evaluation set to compute the ASR at each distortion value.

\noindent\textbf{Blackbox Attack Results Collection.}~For the blackbox attacks, we perform a blackbox attack for each evaluation-set source image, using the set query budgets: 5K, 10K, and 25K. We record the distortion achieved by each source image with a set query budget. To construct ASR vs. distortion, at different distortion thresholds with a given query budget: i)~we compute the number of source images in the evaluation set meeting a given distortion threshold (along the x-axis); and ii)~then divide this by the total number of images in the evaluation set to compute the ASR at each distortion value.

\section{Perturbation Regions and Attack Insights} \label{apd-Salient region and perturbation}

In this section, we provide additional results on the connection between the adversarial perturbations yielded by \rambo and salient regions visualized by the Grad-CAM tool. Effectively, all of the attack methods embedded the target features within the source image where the changes are effectively unnoticeable. However, Fig. \ref{fig:apd-GradCAM-visual} illustrates that a high density of adversarial perturbations yielded by our attack concentrates on a region that is matched to the salient features visualized by the Grad-CAM tool. This is possible because our attack methods employs localized changes to search for adversarial examples and is able to effectively find perturbations targeting salient features of the target class to apply to the input source class image to fool the classifier to classify the source image as the target class.

Further, to help visualize different level of $l_2$ distortions, we include Fig.~\ref{fig:apd-Distortion level visualization}. We illustrate two examples where we showcase the sample adversarial examples crafted by \rambo during the progression of the attack.

\begin{figure}[htp]
    \begin{center}
        \includegraphics[scale=0.25]{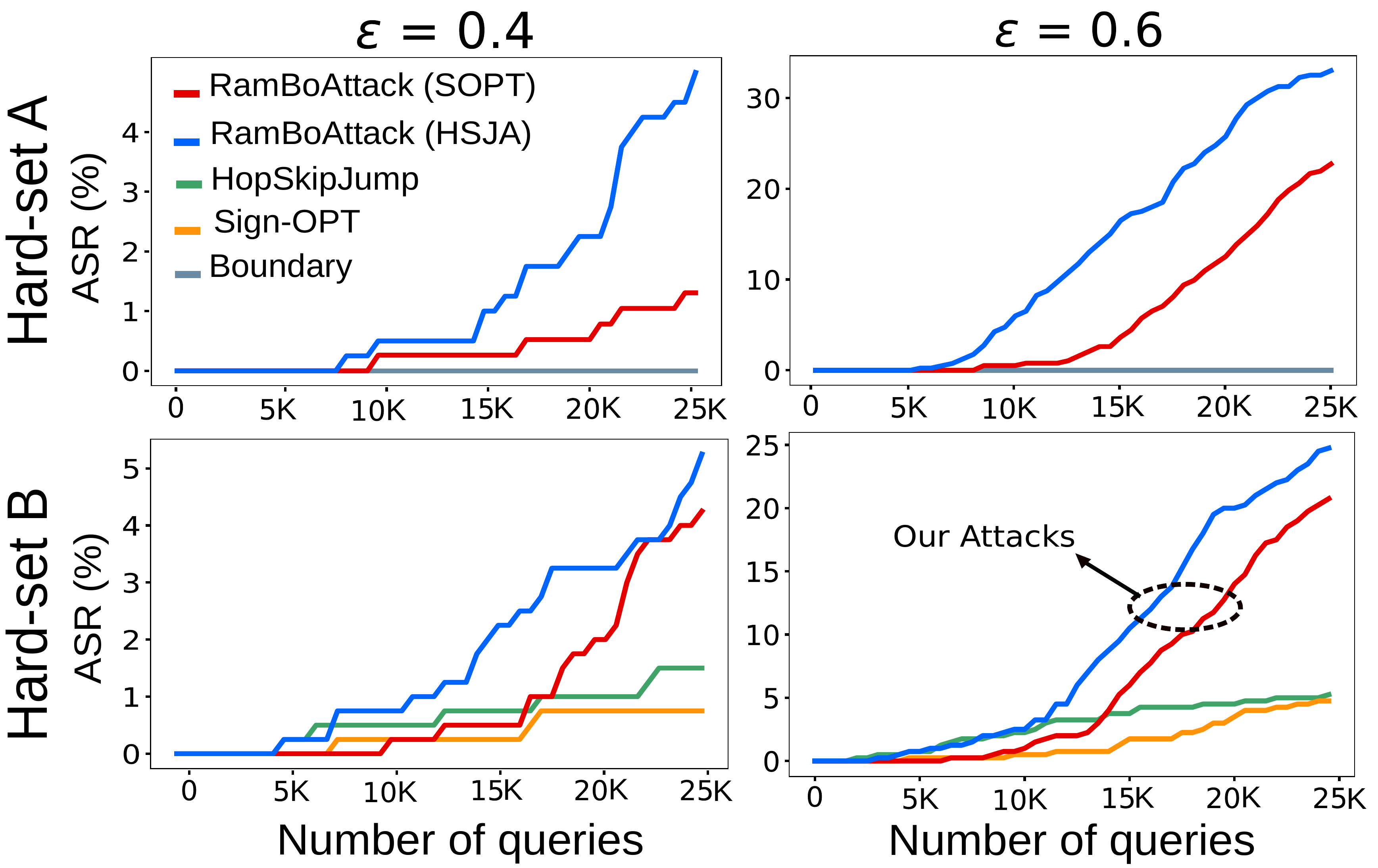}
        \caption{The first row illustrates ASR vs. queries for our \rambos with respect to Boundary attack on hard-set A. The second row illustrates the ASR vs. queries for our \rambos with respect to HopSkipJump and Sign-OPT on hard-set B. The \hardsets are from \texttt{CIFAR10}. For a given query budget, as expected, our \rambos yield similar ASR to Sign-OPT and HSJA with very low query budgets and significantly higher ASR with budgets above 10K queries, where gradient estimation methods do not appear to improve the adversarial example found with increasing numbers of queries.}
        \label{fig:CIFAR10 - ASR vs query different eps}
    \end{center}
    \vspace{-3mm}
\end{figure}

\section{Attack Success Rates vs Query Budgets} \label{apd-The attack success rates vs query number}
In this section, we show results at three different perturbation budgets $\epsilon$ = 0. 4 and 0. 6 for \hardsets A and B from \texttt{CIFAR10} and $\epsilon$  = 10 and 20 for the \hardset selected from \texttt{ImageNet}. The results demonstrate that our attack is significantly more robust than other attacks within 4-11K query budgets. From 11K, \rambo outperforms others. The reason is that, around this region, the gradient estimation method switches to \RSBCD, resulting in much higher attack success rates compared to the baselines. Notably, on the realistic, high-resolution benchmark task \texttt{ImageNet}, \rambo achieves significantly better results compared to the baselines.

\begin{figure}[htp]
    \begin{center}
        \includegraphics[scale=0.25]{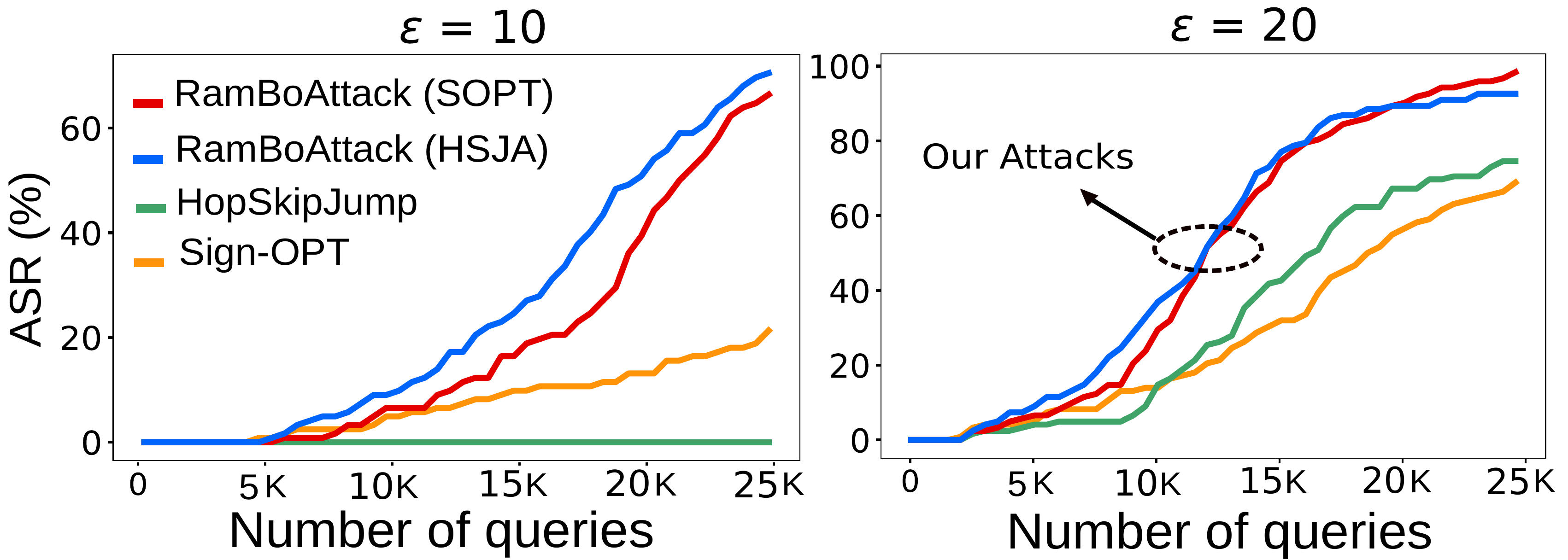}
        \caption{The results illustrate ASR versus queries for our \rambos with respect to HopSkipJump and Sign-OPT on the hard-set from \texttt{ImageNet}. Our \rambos are more query efficient, more robust and are able to yield significantly higher ASR under low distortion settings.}
        \label{fig:ImageNet - ASR vs query different eps}
    \end{center}
    \vspace{-3mm}
\end{figure}

\section{Robustness of \rambo} \label{apd-Robustness of Hybrid Methods}

\begin{figure*}[ht]
    \begin{center}
        \includegraphics[scale=0.5]{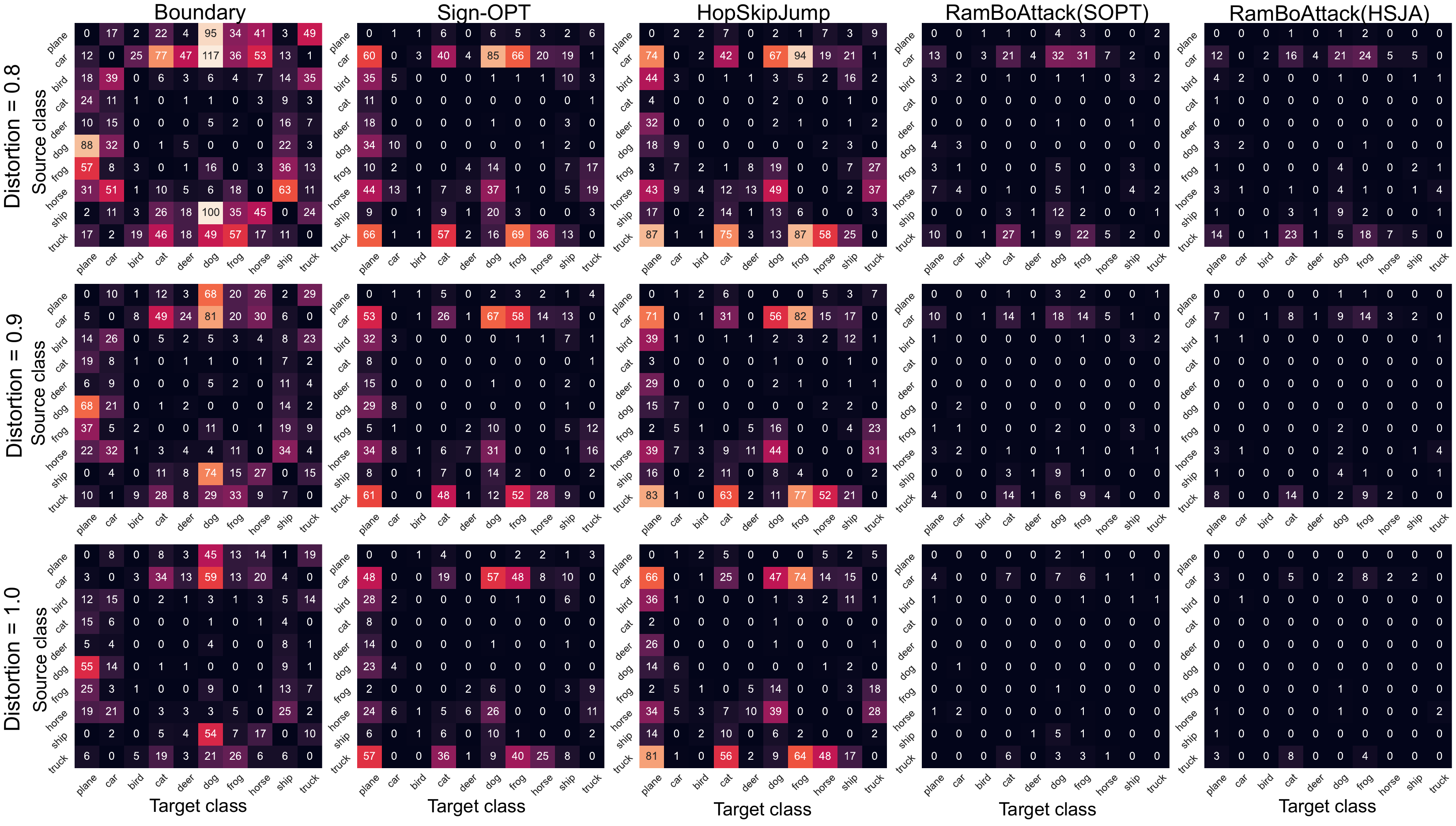}
        \caption{The number of \hard cases on \texttt{CIFAR10} obtained from different attack methods categorized by pairs of source and target classes (at distortion threshold = 0.8, 0.9 and 1.0). \rambos are seen to nearly overcome all of the \hard cases encountered by other decision-based blackbox attack methods; thus, demonstrating the robustness of our proposed attack.}
        \label{fig:adp-Robustness of Hybrid Methods}
    \end{center}
    \vspace{-3mm}
\end{figure*}

Fig. \ref{fig:adp-Robustness of Hybrid Methods} provides further detailed results on \hard cases encountered by different attack methods at distortion threshold of 0.8, 0.9 and 1.0. Compared to Boundary, Sign-OPT and HopSkipJump attacks, our \rambos achieve much lower number of \hard cases at all distortion thresholds.

\end{appendices}

\end{document}